\documentclass{article}
\usepackage{microtype}
\usepackage{graphicx}
\usepackage{booktabs}
\usepackage{hyperref,xurl}

\usepackage[preprint]{icml2026}
\usepackage{amsmath,amssymb,mathtools,amsthm}
\usepackage[capitalize,noabbrev]{cleveref}
\usepackage{enumitem}
\setlist[itemize]{topsep=3pt,itemsep=1pt,parsep=0pt,leftmargin=*}
\setlist[enumerate]{topsep=3pt,itemsep=1pt,parsep=0pt,leftmargin=*}
\usepackage{balance}

\numberwithin{equation}{section}

\AtBeginDocument{%
  \setlength{\abovedisplayskip}{5pt plus 2pt minus 1pt}%
  \setlength{\belowdisplayskip}{5pt plus 2pt minus 1pt}%
  \setlength{\abovedisplayshortskip}{3pt plus 1pt}%
  \setlength{\belowdisplayshortskip}{3pt plus 1pt}%
}
\newtheoremstyle{compactplain}{5pt}{5pt}{\itshape}{}{\bfseries}{.}{.5em}{}
\newtheoremstyle{compactremark}{5pt}{5pt}{}{}{\itshape}{.}{.5em}{}
\theoremstyle{compactplain}
\newtheorem{theorem}{Theorem}[section]
\newtheorem{proposition}[theorem]{Proposition}
\newtheorem{lemma}[theorem]{Lemma}
\newtheorem{corollary}[theorem]{Corollary}
\theoremstyle{definition}

\theoremstyle{compactremark}

\icmltitlerunning{Unbiased Low-Rank Approximation for Repeated Matrix–Vector Multiplication: Stability and Optimal Sampling}
\hypersetup{pdftitle={Unbiased Low-Rank Approximation for Repeated Matrix–Vector Multiplication: Stability and Optimal Sampling},pdfauthor={Kang Liu},pdfsubject={Random matrix dynamics and mean-square stability}}
\begin{document}
\raggedbottom
\twocolumn[
\icmltitle{Unbiased Low-Rank Approximation for Repeated Matrix–Vector Multiplication: Stability and Optimal Sampling}
\begin{icmlauthorlist}
  \icmlauthor{Kang Liu}{xjtu}
\end{icmlauthorlist}
\icmlaffiliation{xjtu}{School of Future Technology, Xi'an Jiaotong University, Xi'an, China}
\icmlcorrespondingauthor{Kang Liu}{kanyo@foxmail.com}
\icmlkeywords{Random matrices, low-rank sampling, mean-square stability, positive operators, Jordan structure}
\vskip 0.3in
]
\printAffiliationsAndNotice{}

\begin{abstract}
Replacing the matrix in an iteration with random low-rank matrices can reduce the cost of matrix–vector multiplication. However, even if the sampled matrices equal the original matrix in expectation, the expected squared norm of the random iterates may grow while the original iteration converges. Under a fixed rank bound, we study how to choose the distribution of these random matrices to minimize this growth when sampling independently at each step. We derive an exact formula for the optimal asymptotic growth factor and use it to determine when the expected squared norm can converge to zero. We then characterize precisely when a fixed sampling distribution achieves this optimal factor and construct finite-support optimal distributions in every attainable case. These results determine the minimum rank required for stability and provide optimal sampling designs under the prescribed rank constraint.
\end{abstract}

\section{Introduction}
\label{sec:introduction}

Many iterative algorithms repeatedly evaluate matrix--vector products, including the power method for eigenvalue computation \citep{lim2017} and the conjugate gradient method for solving linear systems \citep{barrett1994}. For a dense matrix $A\in\mathbb{R}^{d\times d}$, computing $Ax$ costs $O(d^2)$ operations. Replacing $A$ with a rank-$r$ matrix $R$, where $r\ll d$, and evaluating $Rx$ through its low-rank factors reduces this cost to $O(dr)$. Randomization allows us to use a different low-rank matrix at each step while satisfying the unbiasedness condition $\mathbb{E}R=A$, so that the random replacement preserves the action of $A$ in expectation.

Unbiasedness does not ensure stability under iteration. Consider replacing the deterministic update $x_{k+1}=Ax_k$ with $x_{k+1}=R_kx_k$, where the matrices $R_k$ are independent and identically distributed with $\mathbb{E}R_k=A$. Although the mean still satisfies $\mathbb{E}x_k=A^kx_0$, the expected squared norm $\mathbb{E}\|x_k\|_2^2$ can grow even when this mean tends to zero.

For example, let $A=aI_2$ with $0<a<1$. The deterministic iteration multiplies the squared norm by $a^2$ at each step. Now let $R_k$ equal $\operatorname{diag}(2a,0)$ or $\operatorname{diag}(0,2a)$ with equal probability. Every sample has rank one, and $\mathbb{E}R_k=A$, but
\(
\mathbb{E}\|x_{k+1}\|_2^2
=2a^2\,\mathbb{E}\|x_k\|_2^2.
\)
When $1/\sqrt{2}<a<1$, the deterministic iteration converges, whereas the random iteration is unstable in mean square. This amplification comes from the rescaling required for unbiasedness: each coordinate is retained with probability one half, so its contribution must be doubled when selected.

Can different sampling probabilities or candidate matrices avoid this instability? More generally, given $A\in\mathbb{R}^{d\times d}$ and an integer $1\le r\le d$, consider all distributions satisfying
\[
\mathbb{E}R=A,~
\operatorname{rank}R\le r\ \text{a.s.},~
\mathbb{E}\|R\|_F^2<\infty.
\]
We ask two questions: 1) what is the best mean-square growth factor achievable by independent repeated sampling, and 2) does a fixed distribution attain this optimum? We impose no restriction on the sampling basis or the form of the candidate matrices. The rank constraint holds for every sample almost surely, and the distribution does not depend on the initial state.

To determine the optimal growth factor, we first examine how the directions of random outputs affect subsequent iterations. We construct two rank-one sampling distributions with the same mean and the same second-moment matrix $\mathbb{E}[R^\top R]$. Both minimize the Frobenius mean-square approximation error, yet only one produces a mean-square stable iteration. Their expected squared output norms agree after one step, but their outputs are distributed differently across directions and therefore undergo different amplification in subsequent steps. We consequently introduce positive definite quadratic metrics that assign different weights to different directions. For a fixed metric, we establish necessary and sufficient conditions for a prescribed one-step second-moment upper bound to be achievable, together with a corresponding sampling construction.

Optimizing over the metric then yields an exact expression for the best asymptotic mean-square growth factor. Let $q_r(A)$ denote the infimum of this factor over all admissible distributions. We prove that
\[
q_r(A)
=\max\left\{\rho(A)^2,\frac{\mathcal{E}(A)}{r}\right\},
\]
where $\rho(A)$ is the spectral radius and $\mathcal{E}(A):=\sum_{i=1}^{d}|\lambda_i(A)|^2$, with eigenvalues counted according to algebraic multiplicity. The first term is an unavoidable consequence of preserving the mean, and the second captures the additional cost imposed by the rank budget. For the example $A=aI_2$, the formula gives $q_1(A)=2a^2$, showing that the sampling scheme above is already optimal: changing the distribution cannot improve its growth factor. More generally, a mean-square exponentially stable sampling distribution exists if and only if $\rho(A)<1$ and $\mathcal{E}(A)<r$.

To construct sampling distributions that attain this optimum, we further investigate the spectral structure of $A$. We find that the existence of an optimal distribution depends not only on eigenvalue magnitudes but also on the Jordan chains at zero and the structure of the nonzero real eigenvalues and complex conjugate pairs. For example, let $A=\operatorname{diag}(a,J_j(0))$ with $a\ne0$ and $r=1$, where $J_j(0)$ is the size-$j$ Jordan block at zero. The optimal value is always $a^2$, but an optimal distribution exists if and only if $j\le2$. By characterizing the constraints these spectral structures impose on sampling distributions, we establish necessary and sufficient conditions for attainment for every real matrix and rank budget, and construct a finitely supported optimal distribution in every attainable case.

\section{Related work}
\label{sec:related}

Unbiased random approximation represents a deterministic object by a cheaper random one while preserving its mean. \citet{mathe1996} studied the existence of unbiased estimators of linear maps between Banach spaces, with finite expected rank and finite mean squared error uniformly over unit inputs. In finite dimensions, the sampling probabilities can be chosen to minimize approximation error under a budget. \citet{fearnhead2003} minimized a squared error criterion for unbiased resampling in particle filters, and \citet{barnes2024} studied vector sparsification under a cardinality budget for broad classes of convex distortion functions. ATOMO \citep{wang2018} applies the same principle to a specified atomic decomposition, optimizing independent Bernoulli sampling under an expected sparsity budget.

For matrices, \citet[Supplement, Theorem~A.2]{benzing2019} constructed an unbiased approximation with minimum expected squared Frobenius error under an almost sure rank bound. \citet{barnes2025} gave a direct construction by sampling singular components. \citet{bakong2026} used unbiased matrix approximations to reduce the cost of vector--Jacobian products, including objectives weighted by the empirical second moment of the inputs. Such constructions connect to classical sampling and matrix convexity: systematic sampling realizes prescribed inclusion probabilities \citep{madow1949}, while convex combinations of projections describe admissible average subspaces \citep{choi1990,overton1992}. The moment-budget theorem used here combines a prescribed matrix mean with a full matrix upper bound on the second moment.

Directional second moments also govern the effect of compression in optimization. \citet{richtarik2016} optimized probabilities in coordinate descent bounds. \citet{safaryan2021} incorporated smoothness matrices into communication compression, and \citet{flynn2024} analyzed the spectral norm of weighted second moments for coordinate, Haar, and Gaussian sketches. For identity-mean sketches under isotropy and positive semidefiniteness assumptions, \citet{omiya2026} derived the $d/r$ second-moment lower bound in their analysis of randomized subspace acceleration. \citet{li2026} optimized unweighted and weighted trace criteria for positive semidefinite sketches in stochastic gradient estimation. These results motivate treating the direction of the second-moment bound as part of the sampling design. Our moment criterion describes all feasible quadratic upper bounds for arbitrary rectangular means and supplies the construction used in the dynamic optimization.

Repeated sampling leads to products of random matrices. The relation between moment growth, spectral radii, and stochastic Lyapunov functions is developed by \citet{ogura2013,ogura2014}. Fast randomized iteration \citep{lim2017} and randomly sparsified Richardson iteration \citep{weare2026} apply repeated random sparsification to large numerical linear algebra problems. Nonnormal matrix analysis explains how finite-time amplification can coexist with a small asymptotic growth factor \citep{trefethenembree2005}. Within this setting, we optimize the second-moment growth over all distributions with a prescribed matrix mean and an almost sure rank budget. The resulting formula, attaining distributions, and Jordan-structure criterion form the main line of the paper.

\section{Second-moment bounds and sampling}
\label{sec:construction}

To design a sampling law for repeated use, we first seek a second-moment bound that can be applied at every iteration. We then determine which bounds can be realized by unbiased low-rank sampling.

\subsection{Controlling repeated updates}
\label{subsec:quadratic-control}

For a deterministic input $x$, the expected squared norm after one update is $\mathbb E\|Rx\|_2^2=x^\top Hx$, where $H:=\mathbb E[R^\top R]$. After two independent updates,
\[
\mathbb E\|R_1R_0x\|_2^2
=x^\top\mathbb E[R_0^\top H R_0]x.
\]
The second step requires knowing how the random matrix acts on the new quadratic form $H$. We therefore seek a fixed positive definite matrix $P$ for which the quadratic forms before and after an update can be compared directly:
\begin{equation}
\label{eq:quadratic-step-bound}
\mathbb E[R^\top PR]\preceq cP.
\end{equation}
Writing $\|x\|_P^2=x^\top Px$, independence allows us to apply this inequality successively and obtain $\mathbb E\|x_k\|_P^2\le c^k\|x_0\|_P^2$. Comparing the quadratic and Euclidean norms then gives
$\mathbb E\|x_k\|_2^2 \le \kappa_2(P)c^k\|x_0\|_2^2,$
where $\kappa_2(P):=\lambda_{\max}(P)/\lambda_{\min}(P)$. The matrix $P$ weights the directions so that the same one-step comparison remains valid throughout the iteration.

For a fixed sampling law $\mu$, define its asymptotic mean-square growth factor by
\[
g(\mu):=
\lim_{k\to\infty}
\left(
\sup_{\|x_0\|_2=1}
\mathbb E_\mu\|x_k\|_2^2
\right)^{1/k}.
\]
This limit exists. The quadratic bound above gives $g(\mu)\le c$; conversely, every $c>g(\mu)$ admits a positive definite $P$ satisfying \eqref{eq:quadratic-step-bound}. Thus
\[
g(\mu)=
\inf_{\substack{P\succ0,\ c>0\\
\mathbb E_\mu[R^\top PR]\preceq cP}} c.
\]
The proof is given in Appendix~\ref{app:quadratic-growth}. This characterization connects long-term growth to a one-step moment bound and gives a sampling objective: choose the law and the metric to make the feasible factor $c$ as small as possible.

\subsection{The best bound in a fixed metric}
\label{subsec:fixed-metric-bound}

Fix $P\succ0$ and write $B_P:=P^{1/2}AP^{-1/2}$ and $Y:=P^{1/2}RP^{-1/2}$. Similarity preserves the rank constraint, unbiasedness becomes $\mathbb EY=B_P$, and the quadratic bound becomes $\mathbb E[Y^\top Y]\preceq cI$. Controlling samples of $A$ in the metric $P$ is therefore equivalent to controlling samples of $B_P$ in the Euclidean norm.

\begin{theorem}[Feasibility in a fixed metric]
\label{thm:fixed-metric-feasibility}
Given $P\succ0$ and $c\ge0$, there exists a random matrix $R$ with $\mathbb ER=A$ and $\operatorname{rank}(R)\le r$ almost surely such that $\mathbb E[R^\top PR]\preceq cP$ if and only if
$\|B_P\|_2^2\le c, \; \|B_P\|_F^2\le rc.$
Whenever these conditions hold, a law supported on at most $\operatorname{rank}(A)+1$ matrices suffices.
\end{theorem}

Hence the best one-step factor in the fixed metric is
\[
c_r(P):=
\max\left\{
\|B_P\|_2^2,\frac{\|B_P\|_F^2}{r}
\right\}.
\]
The two conditions constrain the largest singular value and the sum of all squared singular values, respectively. The following construction shows how these quantities determine the sampling probabilities. The full feasibility proof is given in Appendix~\ref{app:moment-feasibility}.

\subsection{Deriving the sampling probabilities}
\label{subsec:singular-component-sampling}

If $A=0$, take $R=0$. Otherwise, set $c=c_r(P)>0$, let $s=\operatorname{rank}(A)$, and write the compact singular value decomposition
\[
B_P=\sum_{i=1}^s\sigma_i u_iv_i^\top,
\; \sigma_i>0.
\]
Each term has rank one. We aim to select at most $r$ terms per draw and rescale the selected terms to preserve their mean contributions.

Let $S$ be the random subset of selected components, with marginal probabilities $p_i=\mathbb P(i\in S)>0$. Since component $i$ appears with probability $p_i$, dividing it by $p_i$ when selected preserves its contribution to the mean. Define
\[
Y=\sum_{i\in S}\frac{\sigma_i}{p_i}u_iv_i^\top.
\]
This choice gives $\mathbb EY=B_P$. Orthogonality of the left singular vectors further yields
\[
\mathbb E[Y^\top Y]
=\sum_{i=1}^s\frac{\sigma_i^2}{p_i}v_iv_i^\top.
\]
To bound this matrix by $cI$, each component must satisfy $p_i\ge\sigma_i^2/c$. We choose the smallest admissible marginals,
$p_i=\frac{\sigma_i^2}{c}.$
The definition of $c=c_r(P)$ ensures
\[
0<p_i\le1,
\;
\sum_{i=1}^s p_i=\frac{\|B_P\|_F^2}{c}\le r.
\]
These are precisely the two feasibility conditions: the required probability for each component is at most one, and the total probability does not exceed the rank budget.

For any marginals satisfying these conditions, a coupled subset selection can achieve $\mathbb P(i\in S)=p_i$ while ensuring $|S|\le r$ almost surely. Appendix~\ref{app:moment-feasibility} gives a construction using at most $s+1$ distinct subsets. Finally, setting $R=P^{-1/2}YP^{1/2}$ produces a finite-support law with mean $A$, sample rank at most $r$, and
$\mathbb E[R^\top PR]\preceq c_r(P)P$.

For example, with $A=aI_2$, $r=1$, and $P=I$, we have $c_1(I)=2a^2$. Both components have inclusion probability $1/2$, recovering the two-point sampler from the introduction.

\subsection{Numerical illustration: one-step moments and dynamics}
\label{subsec:one-step-dynamic-illustration}

Take $A=\operatorname{diag}(0.9,0.4)$ and $r=1$. We compare three sampling laws with this mean and rank budget.

The first is the spectral law,
\[
R=
\begin{cases}
\operatorname{diag}(1.3,0),&\text{with probability }9/13,\\
\operatorname{diag}(0,1.3),&\text{with probability }4/13.
\end{cases}
\]
The second is the dense law, which assigns equal probabilities to
\[
R_\pm=
\begin{pmatrix}
0.9&\pm0.6\\
\pm0.6&0.4
\end{pmatrix}.
\]
These laws have the same full right second-moment and error second-moment matrices:
\[
\begin{aligned}
\mathbb E[R^\top R]=\operatorname{diag}(1.17,0.52),~
\mathbb E[(R-A)^\top(R-A)]=0.36I_2.
\end{aligned}
\]
Thus they give the same expected squared output norm and squared approximation error for every deterministic input. Both also attain the minimum Frobenius mean-square error, equal to $0.72$; see Appendix~\ref{app:identical-moments}.

The third law uses the construction above with $P=I$. Since $c_1(I)=\max\{0.81,0.97\}=0.97$, the two singular components have inclusion probabilities $81/97$ and $16/97$. The resulting samples are
\[
R=
\begin{cases}
\operatorname{diag}(97/90,0),&\text{with probability }81/97,\\
\operatorname{diag}(0,97/40),&\text{with probability }16/97.
\end{cases}
\]
This law satisfies $\mathbb E[R^\top R]=0.97I_2$.

Table~\ref{tab:one-step-dynamic-comparison} compares the one-step criteria and the actual growth factors. Both moment columns report the operator norm of the corresponding expected matrix.

\begin{table}[t]
\centering
\caption{Three unbiased rank-one laws. The error and raw moments are
$\|\mathbb E[(R-A)^\top(R-A)]\|_2$ and
$\|\mathbb E[R^\top R]\|_2$, respectively.}
\label{tab:one-step-dynamic-comparison}
\small
\setlength{\tabcolsep}{4pt}
\begin{tabular}{@{}lrrrr@{}}
\toprule
Law & \shortstack{Frobenius\\MSE}
& \shortstack{Error\\moment}
& \shortstack{Raw\\moment}
& $g(\mu)$\\
\midrule
Spectral & $0.72$ & $0.36$ & $1.17$ & $1.17$\\
Dense & $0.72$ & $0.36$ & $1.17$ & $0.97$\\
$P=I$ sampler & $0.97$ & $0.81$ & $0.97$ & $0.97$\\
\bottomrule
\end{tabular}
\end{table}

The spectral law propagates the two coordinate energies separately. The expected squared first coordinate is multiplied by $1.17$ at each step, making the iteration mean-square unstable. The dense law transfers energy between the coordinates and, with $P=A$, satisfies
$\mathbb E[R^\top PR]=0.97P.$
Although this law permits one-step amplification in the Euclidean norm, it contracts at every step in the weighted norm. The third law achieves the same contraction factor directly in the Euclidean norm: for every initial state,
$\mathbb E\|x_k\|_2^2=0.97^k\|x_0\|_2^2.$

\begin{figure}[t]
\centering
\includegraphics[width=0.8\linewidth]{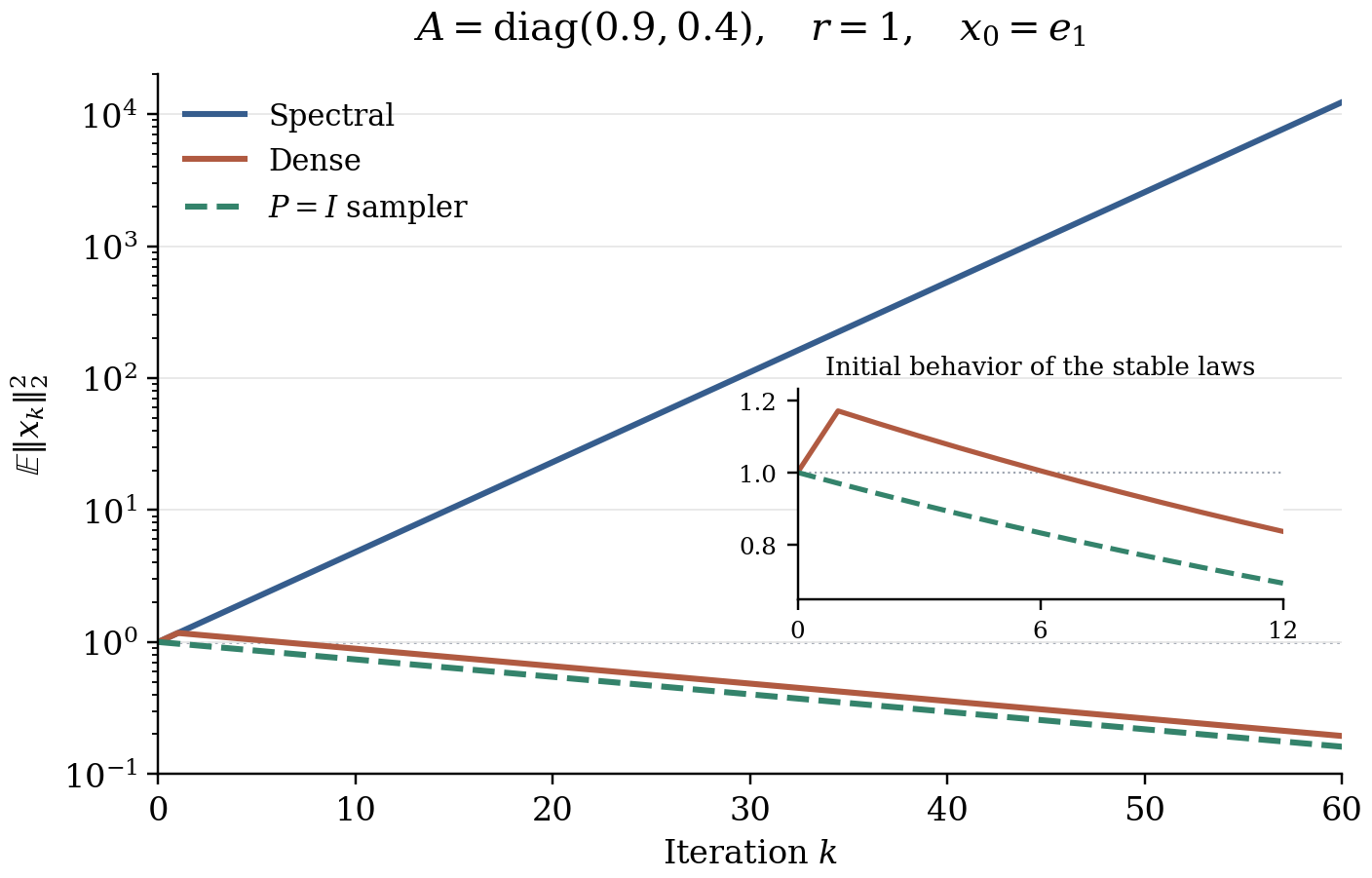}
\caption{Mean-square trajectories of the three sampling laws, computed from their second-moment recursions with $x_0=e_1$. The main plot uses a logarithmic vertical axis; the inset shows the first twelve steps of the two stable laws.}
\label{fig:section3-trajectories}
\end{figure}

Figure~\ref{fig:section3-trajectories} shows the expected squared norms from $x_0=e_1$. The spectral and dense laws both give $1.17$ after the first update, but their subsequent trajectories grow and decay, respectively. The dense law and the $P=I$ sampler have the same asymptotic factor while differing in their initial behavior.

For every fixed metric, we now have the best one-step second-moment bound and a sampling law realizing it. Combining this result with the growth characterization in Section~\ref{subsec:quadratic-control}, we first optimize the law at a fixed metric and then take the infimum over all positive definite metrics. The original problem becomes
$q_r(A)=\inf_{P\succ0}c_r(P).$

\section{Optimal Mean-Square Growth}
\label{sec:optimal-growth}

We now solve the metric optimization problem obtained in the preceding section:
\[
q_r(A)=\inf_{P\succ0}
\max\left\{
\|B_P\|_2^2,\,
\frac{\|B_P\|_F^2}{r}
\right\},
\]
where $B_P=P^{1/2}AP^{-1/2}$. Changing $P$ changes both norms of $B_P$ while preserving its eigenvalues. We first use these eigenvalues to obtain lower bounds, then choose coordinates that approach both bounds simultaneously.

\subsection{Spectral lower bounds}
\label{subsec:spectral-lower-bounds}

Write $\mathcal E(A):=\sum_{i=1}^d|\lambda_i(A)|^2$, with eigenvalues counted according to algebraic multiplicity. Since $B_P$ is similar to $A$, every $P\succ0$ satisfies
$\|B_P\|_2^2\ge\rho(A)^2, \; \|B_P\|_F^2\ge\mathcal E(A).$
The first inequality follows because the operator norm bounds the spectral radius. For the second, a Schur decomposition preserves the Frobenius norm and places the eigenvalues on the diagonal of an upper triangular matrix; its off-diagonal entries can only increase the squared norm. Hence,
\[
q_r(A)\ge
\max\left\{
\rho(A)^2,\,
\frac{\mathcal E(A)}{r}
\right\}.
\]

The two terms reflect different constraints. Preserving the mean requires the random iteration to retain the average action of $A$, giving the lower bound $\rho(A)^2$. Restricting each sample to at most $r$ rank-one components imposes the additional bound $\mathcal E(A)/r$. We next choose coordinates that approach this combined bound.

\subsection{Coordinates that approach both bounds}
\label{subsec:approaching-spectral-bounds}

Choosing a positive definite metric is equivalent to changing coordinates before sampling. Indeed, for any real invertible matrix $S$, set $P=S^\top S$. Its polar decomposition gives $S=UP^{1/2}$ for an orthogonal matrix $U$, and therefore
$SAS^{-1}=UB_PU^\top.$
Thus $SAS^{-1}$ and $B_P$ have the same operator and Frobenius norms. We can consequently search directly for a suitable coordinate transformation $S$.

Consider first
\[
B=
\begin{pmatrix}
\lambda&1\\
0&\lambda
\end{pmatrix},
\; \lambda\in\mathbb R.
\]
Both eigenvalues equal $\lambda$, whereas $\|B\|_F^2=2\lambda^2+1$. The extra term comes from the upper-right entry. For $D_t=\operatorname{diag}(t,1)$ with $t>0$,
\[
D_tBD_t^{-1}=
\begin{pmatrix}
\lambda&t\\
0&\lambda
\end{pmatrix}.
\]
This transformation preserves the eigenvalues and reduces the connecting entry from $1$ to $t$. As $t\to0$,
\[
\|D_tBD_t^{-1}\|_2^2\longrightarrow\lambda^2,\; \|D_tBD_t^{-1}\|_F^2\longrightarrow2\lambda^2.
\]
A single family of coordinate transformations therefore approaches both lower bounds.

The same scaling applies to a general matrix. In real Jordan form, the diagonal spectral blocks are either scalars representing real eigenvalues or two-dimensional blocks
\[
C(a,b)=
\begin{pmatrix}
a&-b\\
b&a
\end{pmatrix}
\]
representing conjugate eigenvalues $a\pm ib$. These blocks are normal. Within each Jordan chain, adjacent spectral blocks are connected by $1$ or $I_2$.

For a chain containing $\ell$ spectral blocks, scale their coordinates successively by $t^{\ell-1},\ldots,t,1$, using the same factor for both coordinates of each two-dimensional block. This leaves the diagonal blocks unchanged and multiplies every connection by $t$. Applying this scaling to all chains gives real invertible matrices $S_t$ such that $S_tAS_t^{-1}\to N$ as $t\to0$, where $N$ is the normal block-diagonal matrix formed by the spectral blocks. Since $N$ retains all eigenvalues of $A$,
\[
\|S_tAS_t^{-1}\|_2^2\longrightarrow\rho(A)^2,\; \|S_tAS_t^{-1}\|_F^2\longrightarrow\mathcal E(A).
\]

Setting $P_t=S_t^\top S_t$ gives positive definite metrics that approach both lower bounds simultaneously. Each $t>0$ yields a valid metric and sampling construction, although the metrics need not converge to a positive definite limit.

\subsection{The optimal factor and stability}
\label{subsec:optimal-factor-stability}

Combining the spectral lower bounds with this coordinate construction gives the exact optimal growth factor.

\begin{theorem}[Optimal mean-square growth]
\label{thm:optimal-mean-square-growth}
For every $A\in\mathbb R^{d\times d}$ and integer $1\le r\le d$,
\begin{equation}
\label{eq:optimal-mean-square-growth}
q_r(A)=
\max\left\{
\rho(A)^2,\,
\frac{\mathcal E(A)}{r}
\right\}.
\end{equation}
For every $c>q_r(A)$, there exist $P\succ0$ and a distribution supported on at most $\operatorname{rank}(A)+1$ matrices such that $\mathbb ER=A$, $\operatorname{rank}R\le r$ almost surely, and
\[
\mathbb E[R^\top PR]\preceq cP.
\]
Independent sampling from this distribution gives, for every deterministic initial state $x_0$,
\begin{equation}
\label{eq:near-optimal-euclidean-bound}
\mathbb E\|x_k\|_2^2
\le \kappa_2(P)c^k\|x_0\|_2^2.
\end{equation}
\end{theorem}

The argument also gives a concrete construction for any requested factor above the optimum. Given $c>q_r(A)$, choose $t>0$ small enough that
\[
\|S_tAS_t^{-1}\|_2^2<c,
\;
\|S_tAS_t^{-1}\|_F^2<rc.
\]
Apply the singular-component construction from Section~\ref{sec:construction} to the transformed matrix, obtaining a random matrix $Y$, and set $R=S_t^{-1}YS_t$. Transforming back preserves unbiasedness and sample rank, while the Euclidean second-moment bound for $Y$ becomes a bound in the metric $P_t=S_t^\top S_t$. The full proof is given in Appendix~\ref{app:optimal-factor}.

Formula~\eqref{eq:optimal-mean-square-growth} describes the effect of increasing the rank budget. As $r$ increases, $\mathcal E(A)/r$ decreases until the requirement to preserve the mean determines the optimal factor through $\rho(A)^2$. For example, $A=aI_d$ gives $q_r(A)=da^2/r$. The rank-one sampler for $A=aI_2$ in the introduction therefore already achieves the optimal factor $2a^2$.

A mean-square exponentially stable sampling distribution exists if and only if
$\rho(A)<1, \; \mathcal E(A)<r.$
When both conditions hold, choosing $q_r(A)<c<1$ in Theorem~\ref{thm:optimal-mean-square-growth} gives exponential decay with a finite-support distribution. Conversely, any mean-square exponentially stable distribution has $g(\mu)<1$, which requires $q_r(A)<1$.

If $A$ is diagonalizable over $\mathbb C$, a fixed real similarity transformation reduces it to a normal block-diagonal matrix with one- and two-dimensional spectral blocks. Both norm lower bounds are then attained, and the construction from Section~\ref{sec:construction} gives a fixed distribution satisfying $g(\mu)=q_r(A)$. For a general matrix, the scaling above gives factors arbitrarily close to the optimum.

\subsection{Numerical illustration: coordinates and iteration}
\label{subsec:coordinate-sampling-experiment}

Consider the matrix family
\[
\begin{gathered}
D=\operatorname{diag}(0.75,0.55,0.35,0.15),\\
S_\eta=I_4+\eta e_1e_2^\top,
\;
A_\eta=S_\eta^{-1}DS_\eta,
\end{gathered}
\]
where $e_i$ denotes the $i$th standard basis vector. The parameter $\eta$ changes the mixing between coordinates while preserving all eigenvalues. Consequently,
$q_r(A_\eta)=\max\{0.5625,\,1.01/r\}$
is independent of $\eta$. The optimal factor is $1.01$ at rank one and decreases to $0.5625$ for every $r\ge2$.

We use $\eta=0,0.25,\ldots,6$ and $r=1,\ldots,4$, comparing three finite-support constructions at each parameter pair, for a total of $300$ distributions. The \emph{Euclidean moment sampler} minimizes $\|\mathbb E[R^\top R]\|_2$ in the original coordinates. The \emph{Frobenius sampler} minimizes $\mathbb E\|R-A_\eta\|_F^2$ in those coordinates. The \emph{transformed sampler} first constructs a Euclidean moment optimum $Y$ for $D$, then sets $R=S_\eta^{-1}YS_\eta$.

The first two constructions use the singular value decomposition of $A_\eta$. The transformed sampler satisfies
\[
\mathbb E[R^\top PR]\preceq q_r(A_\eta)P,
\; P=S_\eta^\top S_\eta,
\]
and therefore attains the optimal mean-square growth factor.

Alongside the growth factor, we record
$a_k:=\sup_{\|x_0\|_2=1}\mathbb E\|x_k\|_2^2,$
the largest expected squared norm at each iteration. The supremum is taken separately for each $k$, so its maximizing initial direction may change with $k$. The curves are obtained by directly propagating second moments using the complete finite supports.

\begin{figure*}[t]
\centering
\includegraphics[width=0.85\textwidth]{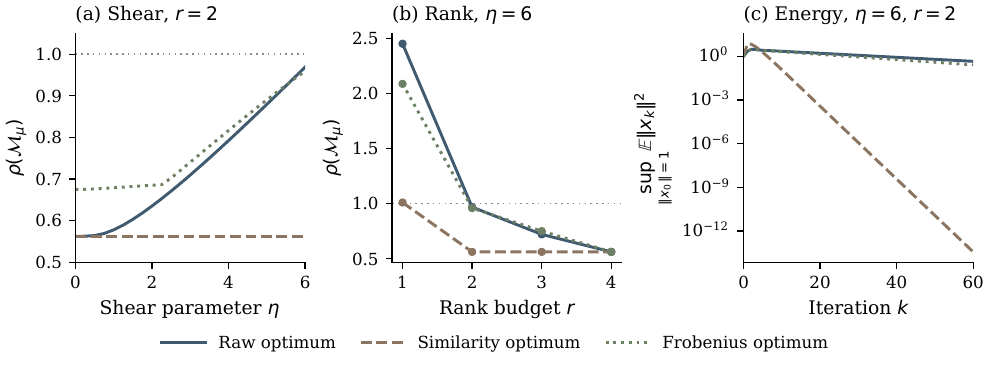}
\caption{Effect of coordinate choice on mean-square growth.
(a) Growth factors as $\eta$ varies, with $r=2$.
(b) Growth factors as $r$ varies, with $\eta=6$.
(c) Worst expected squared norms $a_k$ for $\eta=6$ and $r=2$.
In (a)--(b), the growth factor is labeled $\rho(\mathcal M_\mu)=g(\mu)$, where $\mathcal M_\mu(X)=\mathbb E_\mu[RXR^\top]$ is the second-moment operator; the horizontal dotted line marks the stability boundary $g=1$.
Solid, dotted, and dashed curves correspond to the Euclidean moment, Frobenius, and transformed samplers, respectively.}
\label{fig:coordinate-sampling}
\end{figure*}

At rank two, the growth factors of both constructions in the original coordinates increase with $\eta$ over the tested range. The transformed sampler remains at the theoretical optimum $0.5625$. Table~\ref{tab:coordinate-sampling} gives the values at $\eta=6$.

\begin{table}[t]
\centering
\caption{Growth factors and worst expected squared norms for $\eta=6$ and $r=2$.}
\label{tab:coordinate-sampling}
\small
\setlength{\tabcolsep}{4pt}
\begin{tabular}{@{}lrrr@{}}
\toprule
Sampler & $g(\mu)$ & $a_1$ & $a_{60}$\\
\midrule
Euclidean moment & $0.9687$ & $2.2287$ & $0.4483$\\
Frobenius        & $0.9594$ & $2.2287$ & $0.2568$\\
Transformed      & $0.5625$ & $7.2290$ & $3.76\times10^{-14}$\\
\bottomrule
\end{tabular}
\end{table}

The transformed sampler has greater one-step Euclidean amplification, yet its worst expected squared norm falls to the order of $10^{-14}$ after $60$ iterations. Its one-step bound controls the squared length defined by $P=S_\eta^\top S_\eta$, with the optimal factor $0.5625$. Returning to the Euclidean norm allows greater initial amplification because directions carry different weights, while leaving the asymptotic growth factor unchanged.

This example also illustrates the different roles of $c$ and $\kappa_2(P)$ in \eqref{eq:near-optimal-euclidean-bound}: the former determines the long-term exponential rate, while the latter affects the overall Euclidean bound. For matrices whose connecting entries require increasingly strong scaling, approaching the optimal rate may increase the condition number of the metric. Appendix~C studies the best second-moment bound available under a prescribed condition-number budget.

Each $A_\eta$ in this example is diagonalizable, so a fixed coordinate transformation yields an optimal distribution. To determine when a general matrix admits a fixed distribution attaining $q_r(A)$, we next examine its spectral structure and the allocation of sample rank.

\section{When the Optimum Is Attained}
\label{sec:attainment}

\subsection{A rank-one example}
\label{subsec:rank-one-attainment}

Consider
\[
A_j=\operatorname{diag}(a,J_j(0)),
\; a\ne0,~ r=1,
\]
where $J_j(0)$ denotes the nilpotent Jordan block of size $j$. For every $j$, the optimal factor is $q_1(A_j)=a^2$: the zero eigenvalues contribute nothing to its value. Attainment, however, changes as the zero Jordan chain becomes longer.

First take $j=2$. Let $s$ equal $-1$ or $1$ with equal probabilities, and set $u_s=(1,s,0)^\top$, $v_s=(a,0,s)^\top$, and
\[
R_s=u_sv_s^\top
=
\begin{pmatrix}
a&0&s\\
as&0&1\\
0&0&0
\end{pmatrix}.
\]
Every sample has rank one, and $\mathbb ER_s=A_2$. The random entries have zero mean, while the product of the signs in the two factors produces the required entry $1$ in the zero Jordan block. The zero-chain link thus shares a rank-one factor with the nonzero spectral component.

The same factorization makes repeated sampling easy to analyze. Since $v_s^\top u_t=a$ for every $s,t\in\{-1,1\}$, independent signs $s_0,\ldots,s_{k-1}$ give
\[
R_{s_{k-1}}\cdots R_{s_0}
=
a^{k-1}u_{s_{k-1}}v_{s_0}^\top,
\; k\ge1.
\]
The endpoint vectors have bounded norms, so the asymptotic mean-square growth factor is $a^2$. This fixed two-point distribution attains the optimum.

Replacing $J_2(0)$ by $J_3(0)$ leaves the optimal value unchanged, but no optimal distribution then exists. To explain this length restriction, we examine how an optimal sampler allocates its rank budget.

\subsection{How optimal growth constrains the samples}
\label{subsec:optimal-sample-structure}

We begin with the sampling distribution and identify subspaces invariant under every matrix in its support. Continuing within these subspaces and the corresponding quotients gives coordinates in which all samples share an upper block triangular form:
\[
R=
\begin{pmatrix}
D_1&*&\cdots&*\\
0&D_2&\cdots&*\\
\vdots&\ddots&\ddots&\vdots\\
0&\cdots&0&D_s
\end{pmatrix}.
\]
We refine the decomposition until each diagonal family has no proper nonzero common invariant real subspace. This block structure is determined by the sampling distribution.

The decomposition reduces asymptotic growth to growth on the diagonal blocks. If $\mu_\ell$ denotes the distribution of $D_\ell$, then
$g(\mu)=\max_{1\le\ell\le s}g(\mu_\ell).$
Off-diagonal blocks may cause transient amplification and polynomial prefactors, while the diagonal blocks determine the asymptotic exponential rate. Lemma~\ref{lem:block-radius} in Appendix~\ref{app:block-structure} proves this property.

Now consider the regime in which the rank term determines the optimum, so that $q:=q_r(A)=\mathcal E(A)/r>0$. Suppose a distribution attains $g(\mu)=q$, and write $B_\ell=\mathbb ED_\ell$. For each irreducible diagonal family, there is a positive definite matrix $P_\ell$ such that $\mathbb E[D_\ell^\top P_\ell D_\ell]=g(\mu_\ell)P_\ell$; see Lemma~\ref{lem:perron}. This identity expresses its asymptotic growth as an exact one-step second-moment relation. Since $g(\mu_\ell)\le q$, applying the rank--moment inequality in this metric gives
\[
\begin{aligned}
r
\ge \mathbb E\operatorname{rank}R
\ge \sum_{\ell=1}^s
       \mathbb E\operatorname{rank}D_\ell
\ge \frac1q\sum_{\ell=1}^s\mathcal E(B_\ell)
=\frac{\mathcal E(A)}q
=r.
\end{aligned}
\]
Every inequality must therefore be an equality. The expected rank required by the nonzero spectrum exhausts the entire budget, so the remaining parts of the samples must share rank with these diagonal blocks.

The equality conditions impose a specific structure. Every nonzero diagonal mean $B_\ell$ is invertible, whereas a diagonal block with zero mean vanishes in every sample. Moreover, merging consecutive zero-mean diagonal blocks produces a block that also vanishes in every sample. If there are $k$ nonzero diagonal means, they consequently separate at most $k+1$ such zero layers, including the layers at the two ends.

Let $\nu_0(A)$ denote the largest Jordan block size at zero, with $\nu_0(A)=0$ when $A$ is invertible. In the block expansion of the resolvent $(zI-A)^{-1}$ near zero, each zero layer contributes at most one factor $z^{-1}$ along a block path, while invertible diagonal blocks contribute factors analytic at zero. The pole order at zero equals the largest zero Jordan block size. Thus, as detailed in Appendix~\ref{app:rank-equality}, this interleaved block form requires
\begin{equation}
\label{eq:zero-layer-bound}
\nu_0(A)\le k+1.
\end{equation}
Thus the maximum zero-chain length is controlled by the number of nonzero blocks available to separate the zero layers.

\subsection{The complete attainment criterion}
\label{subsec:complete-attainment-criterion}

To express the number of available nonzero blocks in terms of $A$, let $b(A)$ count its nonzero real eigenvalues and nonreal conjugate pairs with algebraic multiplicity, counting each pair once. Equivalently, $b(A)$ is the number of nonzero one- or two-dimensional diagonal spectral blocks in a real Jordan form. For example, a nonzero real Jordan chain of length $\ell$ contributes $\ell$ such blocks.

Each nonzero diagonal mean in the preceding decomposition contains at least one real spectral block, so $k\le b(A)$. Combining this inequality with \eqref{eq:zero-layer-bound} gives $\nu_0(A)\le b(A)+1$. This necessary condition also suffices to construct an optimal distribution.

\begin{theorem}[Attainment of the optimum]
\label{thm:attainment}
For every $A\in\mathbb R^{d\times d}$ and integer $1\le r\le d$, attainment of $q_r(A)$ is characterized as follows.
\begin{enumerate}
\setlength{\itemsep}{0pt}
\item If $\rho(A)=0$, the optimum is attained.
\item If $\rho(A)^2>\mathcal E(A)/r$, the optimum is attained.
\item If $\mathcal E(A)/r\ge\rho(A)^2>0$, the optimum is attained if and only if
$\nu_0(A)\le b(A)+1.$
\end{enumerate}
In every attainable case, a finite-support distribution attains the optimum.
\end{theorem}

Every invertible real matrix therefore admits an optimal distribution, including matrices with nontrivial Jordan chains. For the opening example $A_j=\operatorname{diag}(a,J_j(0))$, we have $b(A_j)=1$, so a rank-one optimal distribution exists exactly when $j\le2$.

The distinction between real eigenvalues and conjugate pairs also affects attainment. Consider
\[
\begin{aligned}
A_{\mathrm{real}}
&=\operatorname{diag}(a,a,J_3(0)),\\
A_{\mathrm{pair}}
&=\operatorname{diag}(C_a,J_3(0)),
\end{aligned}
\;
C_a=
\begin{pmatrix}
0&-a\\
a&0
\end{pmatrix},
\]
where $a\ne0$. These matrices have the same eigenvalue moduli and the same zero Jordan block, and both have rank-one optimal value $2a^2$. Yet $b(A_{\mathrm{real}})=2$ and $b(A_{\mathrm{pair}})=1$, so the first admits an optimal distribution and the second does not. Under real sampling, a conjugate pair supplies one two-dimensional real spectral block while contributing two squared eigenvalue moduli to $\mathcal E(A)$.

\subsection{Constructing finite-support optimal distributions}
\label{subsec:optimal-finite-support}

We first address the regime governed by the chain-length criterion:
\[
q=\frac{\mathcal E(A)}r\ge\rho(A)^2>0,
\;
\nu_0(A)\le b(A)+1.
\]
The construction follows the idea of Section~\ref{subsec:rank-one-attainment}: allocate rank to the nonzero spectrum, then use the same rank-one factors to realize the zero-chain links.

In real Jordan coordinates, let $C_1,\ldots,C_b$ be the diagonal spectral blocks of the nonzero part, where $b=b(A)$. Each block has dimension $h_j\in\{1,2\}$ and satisfies
$C_j^\top C_j=\omega_j^2I_{h_j}, \; \omega_j>0.$
Assign each coordinate in block $j$ the selection probability $p_j=\omega_j^2/q$. These probabilities satisfy
\[
0<p_j\le1,
\;
\sum_{j=1}^b h_jp_j
=\frac{\mathcal E(A)}q
=r.
\]
A coupled subset can therefore realize these marginals while selecting exactly $r$ coordinates on every draw.

Next, place the coordinates of each zero Jordan chain into spaces $H_0,\ldots,H_b$ according to their positions along the chain. Order the full coordinates as
\begin{equation}
\label{eq:attainment-interleaved-order}
H_0,\ C_1,\ H_1,\ C_2,\ \ldots,\ C_b,\ H_b.
\end{equation}
The condition $\nu_0(A)\le b+1$ ensures that every zero chain fits within these layers. Different chains can share a layer, so the required number of layers depends on the longest chain.

To incorporate the links between zero layers into existing rank-one factors, we use centered random vectors. The basic identity is that, for fixed $u,v,L,M$ and a random vector satisfying $\mathbb E\xi=0$ and $\mathbb E[\xi\xi^\top]=I$,
\[
\mathbb E\bigl[(u+L\xi)(v+M\xi)^\top\bigr]
=uv^\top+LM^\top.
\]
Every realization has rank at most one, while the product of the correlated random terms creates a prescribed link in the mean. Placing $L\xi$ and $M\xi$ in adjacent zero layers and choosing their coefficients appropriately realizes the zero-chain links. The random-sign construction in Section~\ref{subsec:rank-one-attainment} is the simplest instance of this mechanism. These random vectors are chosen independently of the sampled subset. Jordan links between nonzero spectral blocks are included in the rank-one decompositions of the corresponding rows.

The resulting sample is a sum of $r$ rank-one factors and is upper block triangular in the order \eqref{eq:attainment-interleaved-order}. Its diagonal blocks on the zero layers vanish. On each nonzero spectral block, the random diagonal block has the form
$D_j=p_j^{-1}\Pi_jC_j,$
where $\Pi_j$ is the diagonal projection onto the selected coordinates and $\mathbb E\Pi_j=p_jI_{h_j}$. Hence,
\[
\mathbb E[D_j^\top D_j]
=p_j^{-1}C_j^\top C_j
=qI_{h_j}.
\]
Every diagonal block has growth factor at most $q$, and the full distribution therefore attains $q_r(A)=q$. Both the coupled subset and the random signs can have finite support. Appendix~\ref{app:attainment} gives the complete factor formulas.

The other two regimes admit more direct constructions. If $\rho(A)=0$, transform $A$ to strictly upper triangular form. On each draw, select one nonzero entry, divide it by its selection probability, and set all other entries to zero. This gives an unbiased rank-one sampler. Every sample is strictly upper triangular, so every product of $d$ samples vanishes and the growth factor is zero. For $A=0$, take $R=0$.

If $\rho(A)^2>\mathcal E(A)/r$, use real coordinates in which $A=\operatorname{diag}(A_{\mathrm p},A_{\mathrm s})$. The block $A_{\mathrm p}$ contains all eigenvalues of modulus $\rho(A)$ together with their full Jordan chains, and $A_{\mathrm s}$ contains the remaining spectrum. If $m$ is the dimension of $A_{\mathrm p}$, then
$m<r, \; q_{r-m}(A_{\mathrm s})<\rho(A)^2.$
Keep $A_{\mathrm p}$ deterministic and apply the construction of Section~\ref{sec:optimal-growth} to obtain a sampler $R_{\mathrm s}$ for $A_{\mathrm s}$ with growth factor below $\rho(A)^2$. Then
$R=\operatorname{diag}(A_{\mathrm p},R_{\mathrm s})$
respects the rank budget and has growth factor exactly $\rho(A)^2$.

\subsection{Numerical illustrations of attainment and approximation}
\label{subsec:attainment-illustrations}

We first consider eight finite-support constructions. They cover real spectral blocks, conjugate pairs, nontrivial Jordan chains, multiple zero chains, and all three regimes of Theorem~\ref{thm:attainment}. For compact notation, write
\[
C=
\begin{pmatrix}
1&-1\\
1&1
\end{pmatrix},
\;
B=
\begin{pmatrix}
1&3\\
0&1
\end{pmatrix},
\;
H=
\begin{pmatrix}
C&I_2\\
0&C
\end{pmatrix}.
\]
Table~\ref{tab:attainment-illustrations} lists the means up to similarity, with $N$ denoting the number of entries in the finite sampling list. Every construction has the prescribed mean and respects the rank budget on every draw. Its growth factor is obtained from the characteristic polynomial of the second-moment operator.

\begin{table*}[t]
\centering
\caption{Finite-support optimal distributions covering the three attainment regimes. In every row, the constructed distribution satisfies $\mathbb ER=A$, $\operatorname{rank}R\le r$ almost surely, and $g(\mu)=q_r(A)$. The column $N$ counts entries in the finite sampling list.}
\label{tab:attainment-illustrations}
\small
\setlength{\tabcolsep}{8pt}
\begin{tabular}{@{}lrrrrr@{}}
\toprule
Mean $A$, up to similarity
& $r$ & $b(A)$ & $\nu_0(A)$ & $N$ & $q_r(A)=g(\mu)$\\
\midrule
$\operatorname{diag}(1,2,J_3(0))$
& 1 & 2 & 3 & 4 & 5\\
$\operatorname{diag}(C,J_2(0))$
& 1 & 1 & 2 & 4 & 4\\
$\operatorname{diag}(1,1,J_3(0))$
& 2 & 2 & 3 & 4 & 1\\
$\operatorname{diag}(B,J_3(0))$
& 1 & 2 & 3 & 4 & 2\\
$\operatorname{diag}(H,J_3(0))$
& 2 & 2 & 3 & 8 & 4\\
$\operatorname{diag}(1,1,J_3(0),J_2(0))$
& 2 & 2 & 3 & 8 & 1\\
$\operatorname{diag}(1,\tfrac12,J_4(0))$
& 2 & 2 & 4 & 6 & 1\\
$J_4(0)$
& 1 & 0 & 4 & 4 & 0\\
\bottomrule
\end{tabular}
\end{table*}

The first six examples attain the optimum by interleaving zero layers with nonzero spectral blocks. The seventh satisfies $\rho(A)^2>\mathcal E(A)/r$, so an optimal distribution exists even though its zero-chain length exceeds $b(A)+1$. In the final example, products vanish after finitely many steps, giving growth factor zero. The sampling construction and complete characteristic polynomials are given in Appendices~\ref{app:attainment-constructions} and~\ref{app:attainment-examples}, respectively.

Now consider the unattainable case
$r=1,~A=\operatorname{diag}(0.6,J_3(0)).$
Here $q_1(A)=0.36$, but $\nu_0(A)=3>b(A)+1=2$, so no fixed distribution attains $0.36$.

An explicit family illustrates how this value can be approached. For $\varepsilon>0$, set $c_\varepsilon=0.36+2\varepsilon^2$. Among the three nonzero rows of $A$, select the row containing $0.6$ with probability $0.36/c_\varepsilon$ and each of the two rows containing a zero-chain link with probability $\varepsilon^2/c_\varepsilon$. Retain only the selected row and divide it by its selection probability. The resulting distribution is unbiased, and every sample has rank one.

Its growth factor and sample second moment are
\[
g(\mu_\varepsilon)=c_\varepsilon,\; \mathbb E\|R\|_F^2
=c_\varepsilon+\frac{2c_\varepsilon}{\varepsilon^2}.
\]
Table~\ref{tab:approaching-unattained-value} gives the values at six scales.

\begin{table}[t]
\centering
\caption{A three-point sampling family approaching the unattained optimal value $0.36$. The gap is $g(\mu_\varepsilon)-0.36$.}
\label{tab:approaching-unattained-value}
\small
\setlength{\tabcolsep}{5pt}
\begin{tabular}{@{}rrrr@{}}
\toprule
$\varepsilon$ & $g(\mu_\varepsilon)$ & Gap
& $\mathbb E\|R\|_F^2$\\
\midrule
0.40 & 0.6800 & 0.3200 & 9.1800\\
0.20 & 0.4400 & 0.0800 & 22.4400\\
0.10 & 0.3800 & 0.0200 & 76.3800\\
0.05 & 0.3650 & 0.0050 & 292.3650\\
0.02 & 0.3608 & 0.0008 & 1804.3608\\
0.01 & 0.3602 & 0.0002 & 7204.3602\\
\bottomrule
\end{tabular}
\end{table}

As $\varepsilon$ decreases, the zero-chain rows are selected less often and receive larger compensating weights. Within this family, the gap to the optimal growth factor is $2\varepsilon^2$, while the sample second moment grows on the order of $\varepsilon^{-2}$. Every positive $\varepsilon$ gives an admissible distribution with finite second moment.

For a given $A$ and $r$, the formula in Section~\ref{sec:optimal-growth} determines the best mean-square growth factor, and the spectral criterion in this section determines whether a fixed distribution realizes that value. In every attainable case, the constructions above give a finite-support optimal distribution. When the optimum is unattained, Section~\ref{sec:optimal-growth} still provides a finite-support sampler with a quadratic bound for every prescribed factor $c>q_r(A)$.

\section{Conclusion and Discussion}
\label{sec:conclusion}

We characterized the optimal asymptotic mean-square growth of independently repeated unbiased low-rank matrix sampling. An exact feasibility criterion for quadratic second-moment bounds yields the optimal growth factor and a necessary and sufficient condition for mean-square exponential stability. We further characterized when a fixed distribution attains this optimum through the interaction between zero Jordan chains and nonzero real spectral blocks, and constructed finite-support optimal distributions in every attainable case.

Two directions remain for sampling design. First, finite-horizon performance depends on transient amplification as well as asymptotic growth. Our examples show that a smaller growth factor can accompany greater initial amplification, and that approaching an unattained optimum can involve increasing sample second moments. The sharp conditioning bounds in Appendix~\ref{app:extensions} quantify one aspect of this tradeoff and motivate optimizing finite-horizon performance under conditioning or moment constraints. Second, temporal dependence changes the achievable growth rate: independent time-varying laws cannot improve the optimum, whereas sampling with memory can improve it even under conditional unbiasedness. Characterizing optimal sampling policies under prescribed information and rank constraints remains an open direction.

\bibliographystyle{icml2026}
\bibliography{reference}

\clearpage
\onecolumn
\appendix
\numberwithin{theorem}{section}
\numberwithin{equation}{section}
\numberwithin{table}{section}
\numberwithin{figure}{section}
\section*{Appendix contents}
\begingroup
\small
\setlength{\tabcolsep}{0pt}
\renewcommand{\arraystretch}{1.08}
\noindent\begin{tabular*}{\textwidth}{@{\extracolsep{\fill}}lr}
\hyperref[app-toc-a]{\textbf{A.~ Proofs of the core results}} & \pageref{app-toc-a} \\
\hyperref[app-toc-a-1]{\hspace*{1em}A.1~ Second-moment propagation and quadratic bounds} & \pageref{app-toc-a-1} \\
\hyperref[app-toc-a-2]{\hspace*{1em}A.2~ Mean, rank, and second moment} & \pageref{app-toc-a-2} \\
\hyperref[app-toc-a-3]{\hspace*{1em}A.3~ Finite subset sampling with prescribed marginals} & \pageref{app-toc-a-3} \\
\hyperref[app-toc-a-4]{\hspace*{1em}A.4~ Feasibility of general quadratic moment bounds} & \pageref{app-toc-a-4} \\
\hyperref[app-toc-a-5]{\hspace*{1em}A.5~ The optimal growth factor and the stability threshold} & \pageref{app-toc-a-5} \\
\hyperref[app-toc-b]{\textbf{B.~ Attainment and construction of optimal distributions}} & \pageref{app-toc-b} \\
\hyperref[app-toc-b-1]{\hspace*{1em}B.1~ Common block structure and positive definite eigenmatrices} & \pageref{app-toc-b-1} \\
\hyperref[app-toc-b-2]{\hspace*{1em}B.2~ Rank equality and the length of zero Jordan chains} & \pageref{app-toc-b-2} \\
\hyperref[app-toc-b-3]{\hspace*{1em}B.3~ Finite-support constructions in all three regimes} & \pageref{app-toc-b-3} \\
\hyperref[app-toc-b-4]{\hspace*{1em}B.4~ Optimal positive definite quadratic bounds} & \pageref{app-toc-b-4} \\
\hyperref[app-toc-b-5]{\hspace*{1em}B.5~ Rank-one structure and a uniform bound with one nonzero eigenvalue} & \pageref{app-toc-b-5} \\
\hyperref[app-toc-b-6]{\hspace*{1em}B.6~ Complete numerical records for attainable and unattainable cases} & \pageref{app-toc-b-6} \\
\hyperref[app-toc-c]{\textbf{C.~ One-step error, finite-time bounds, and temporal dependence}} & \pageref{app-toc-c} \\
\hyperref[app-toc-c-1]{\hspace*{1em}C.1~ One-step moments, approximation error, and optimal laws} & \pageref{app-toc-c-1} \\
\hyperref[app-toc-c-2]{\hspace*{1em}C.2~ One-step criteria and repeated sampling} & \pageref{app-toc-c-2} \\
\hyperref[app-toc-c-3]{\hspace*{1em}C.3~ Conditioning and finite-time bounds} & \pageref{app-toc-c-3} \\
\hyperref[app-toc-c-4]{\hspace*{1em}C.4~ Independent schedules and sampling with memory} & \pageref{app-toc-c-4} \\
\hyperref[app-toc-c-5]{\hspace*{1em}C.5~ Computation and reproducibility} & \pageref{app-toc-c-5} \\
\end{tabular*}
\endgroup
\medskip

\section{Proofs of the core results}\label{app-toc-a}
\label{app:core}
\label{app:dynamics}

This appendix follows the analysis in Sections~3 and~4 of the main text. We first express mean-square growth of independent random products through the spectral radius of the second-moment operator and show how positive definite quadratic forms characterize this growth. We then prove an inequality relating the mean, rank, and second moment, and construct random subsets with prescribed marginals. Together, these results give an exact feasibility criterion for general quadratic moment bounds. Finally, we optimize the metric to establish the optimal growth factor and the stability threshold.

\subsection{Second-moment propagation and quadratic bounds}\label{app-toc-a-1}
\label{app:quadratic-growth}

We establish the growth characterization used in the main text and explain why optimization over positive definite quadratic forms gives the exact asymptotic factor of a fixed distribution.

For $A\in\mathbb R^{d\times d}$ and an integer $1\le r\le d$, define the admissible class
\[
\mathcal C_r(A):=\left\{\mu:\mathbb E_\mu R=A,\;\operatorname{rank}R\le r\ \text{almost surely},\;\mathbb E_\mu\|R\|_F^2<\infty\right\}.
\]
Let $R_0,R_1,\ldots$ be independent with common distribution $\mu\in\mathcal C_r(A)$, and set $\Phi_0=I_d$ and $\Phi_k=R_{k-1}\cdots R_0$. The iteration from a deterministic state $x_0$ then satisfies $x_k=\Phi_kx_0$. Define its mean-square growth factor by $g(\mu):=\lim_{k\to\infty}(\sup_{\|x\|_2=1}\mathbb E\|\Phi_kx\|_2^2)^{1/k}$; the existence of this limit is proved below. The optimal value is $q_r(A):=\inf_{\mu\in\mathcal C_r(A)}g(\mu)$.

Write $\operatorname{Sym}^d$ for the space of real symmetric matrices. The second-moment operator and its adjoint with respect to the trace inner product are $\mathcal M_\mu(X)=\mathbb E[RXR^\top], \; \mathcal M_\mu^*(P)=\mathbb E[R^\top PR]$. Both maps preserve positive semidefiniteness. The finite second-moment assumption makes them bounded linear maps on this finite-dimensional space. For $P\succ0$, write $\|x\|_P^2=x^\top Px$ and $\kappa_2(P)=\lambda_{\max}(P)/\lambda_{\min}(P)$. We also write $\rho(A)$ for the spectral radius and $\mathcal E(A)=\sum_{i=1}^d|\lambda_i(A)|^2$, with eigenvalues counted according to algebraic multiplicity.

\begin{lemma}[Mean-square growth and quadratic bounds]
\label{lem:lyapunov}
For every $\mu\in\mathcal C_r(A)$,
\begin{equation}
\begin{aligned}
g(\mu)
&=\lim_{k\to\infty}
\left(\sup_{\|x\|_2=1}\mathbb E\|\Phi_kx\|_2^2\right)^{1/k}
=\rho(\mathcal M_\mu),\\
\rho(\mathcal M_\mu)
&=\inf_{P\succ0}
\lambda_{\max}\!\left(
P^{-1/2}\mathcal M_\mu^*(P)P^{-1/2}\right).
\end{aligned}
\label{eq:app-a-growth-quadratic}
\end{equation}
If $\mathcal M_\mu^*(P)\preceq cP$ for some $P\succ0$ and $c\ge0$, then, for every $x\in\mathbb R^d$ and $k\ge1$,
\begin{equation}
\mathbb E\|\Phi_kx\|_2^2
\le\kappa_2(P)c^k\|x\|_2^2.
\label{eq:app-a-euclidean-bound}
\end{equation}
Conversely, every $c>\rho(\mathcal M_\mu)$ admits a $P\succ0$ such that $\mathcal M_\mu^*(P)\prec cP$.
\end{lemma}

\begin{proof}
Independence and successive conditioning give $a_k:=\sup_{\|x\|_2=1}\mathbb E\|\Phi_kx\|_2^2 =\bigl\|(\mathcal M_\mu^*)^k(I_d)\bigr\|_2$. For any positive linear map $T:\operatorname{Sym}^d\to\operatorname{Sym}^d$, the norm induced by the matrix operator norm satisfies $\|T\|_{2\to2}=\|T(I_d)\|_2$. Indeed, $\|X\|_2\le1$ implies $-I_d\preceq X\preceq I_d$. Positivity then gives $-T(I_d)\preceq T(X)\preceq T(I_d)$, so $\|T(X)\|_2\le\|T(I_d)\|_2$, with equality at $X=I_d$. Applying this identity to $(\mathcal M_\mu^*)^k$ and using Gelfand's formula \citep[Section~5.6]{hornjohnson2013} yields $\lim_{k\to\infty}a_k^{1/k} =\rho(\mathcal M_\mu^*) =\rho(\mathcal M_\mu)$.

Now suppose $\mathcal M_\mu^*(P)\preceq cP$. Independence of $R_k$ and $x_k$ gives $\mathbb E[x_{k+1}^\top Px_{k+1}\mid x_k] \le c\,x_k^\top Px_k$. Iterating this inequality yields $\mathbb E\|\Phi_kx\|_P^2\le c^k\|x\|_P^2$. The inequalities $\lambda_{\min}(P)\|z\|_2^2\le\|z\|_P^2\le\lambda_{\max}(P)\|z\|_2^2$ then give \eqref{eq:app-a-euclidean-bound}. Taking $k$th roots and passing to the limit shows that $\rho(\mathcal M_\mu)\le c$. If $c=0$, the same conclusion follows directly, since $\mathbb E[R^\top PR]=0$ forces $R=0$ almost surely.

For the converse, fix $c>\rho(\mathcal M_\mu)$ and define
\begin{equation}
P=\sum_{k=0}^{\infty}c^{-k-1}(\mathcal M_\mu^*)^k(I_d).
\label{eq:app-a-resolvent}
\end{equation}
Since $\rho(\mathcal M_\mu^*/c)<1$, this series converges in norm. Every term is positive semidefinite and the first is $c^{-1}I_d$, so $P\succ0$. Shifting the series gives $\mathcal M_\mu^*(P)=cP-I_d\prec cP$. Every positive definite $P$ gives an upper bound on the spectral radius, and \eqref{eq:app-a-resolvent} makes these bounds arbitrarily close to that radius. This proves the second equality in \eqref{eq:app-a-growth-quadratic}.
\end{proof}

Thus a fixed distribution is mean-square exponentially stable if and only if $\rho(\mathcal M_\mu)<1$. A spectral radius below one permits the choice $\rho(\mathcal M_\mu)<c<1$ in the lemma. Conversely, any uniform bound $\mathbb E\|\Phi_kx\|_2^2\le Cc^k\|x\|_2^2$ with $c<1$ implies a spectral radius below one by \eqref{eq:app-a-growth-quadratic}. The lemma also establishes the quadratic optimization characterization for a fixed distribution used in the main text.

\subsection{Mean, rank, and second moment}\label{app-toc-a-2}
\label{app:rank-moment}

The rank constraint enters the second-moment bound through the range of each sample. The following inequality expresses this geometric restriction as a matrix inequality for the mean.

\begin{lemma}[Mean, rank, and second moment]
\label{lem:rank}
Let $Y\in\mathbb R^{m\times n}$ have finite second moment, set $D=\mathbb EY$, and let $\Pi_Y$ be the orthogonal projection onto $\operatorname{range}(Y)$. If $\mathbb E[Y^\top Y]\preceq\gamma I_n$ for some $\gamma>0$, then
\begin{equation}
DD^\top\preceq\gamma\mathbb E\Pi_Y.
\label{eq:app-a-rank-matrix}
\end{equation}
In particular,
\begin{equation}
\|\mathbb EY\|_F^2
\le\mathbb E[\operatorname{rank}Y]\,
\bigl\|\mathbb E[Y^\top Y]\bigr\|_2.
\label{eq:app-a-rank-trace}
\end{equation}
\end{lemma}

\begin{proof}
The range projection is $\Pi_Y=YY^\dagger$, where $\dagger$ denotes the Moore--Penrose inverse \citep{penrose1955}. The representation $\Pi_Y=\lim_{t\downarrow0} Y(Y^\top Y+tI_n)^{-1}Y^\top$ shows that $\Pi_Y$ is measurable in $Y$. Its expectation exists because $\|\Pi_Y\|_2\le1$. Since $\Pi_YY=Y$ and $\Pi_Y^2=\Pi_Y$,
\[
\begin{pmatrix}
Y^\top Y&Y^\top\\
Y&\Pi_Y
\end{pmatrix}
=
\begin{pmatrix}Y^\top\\\Pi_Y\end{pmatrix}
\begin{pmatrix}Y&\Pi_Y\end{pmatrix}
\succeq0.
\]
Taking expectations and increasing the upper-left block to $\gamma I_n$ gives
\[
\begin{pmatrix}
\gamma I_n&D^\top\\
D&\mathbb E\Pi_Y
\end{pmatrix}\succeq0.
\]
The Schur complement of the upper-left block \citep{hornzhang2005} yields $\mathbb E\Pi_Y-\gamma^{-1}DD^\top\succeq0$, proving \eqref{eq:app-a-rank-matrix}. If $\gamma=\|\mathbb E[Y^\top Y]\|_2>0$, taking traces and using $\operatorname{tr}\Pi_Y=\operatorname{rank}Y$ gives \eqref{eq:app-a-rank-trace}. If this norm vanishes, then $Y=0$ almost surely and \eqref{eq:app-a-rank-trace} still holds.
\end{proof}

\subsection{Finite subset sampling with prescribed marginals}\label{app-toc-a-3}
\label{app:subset-sampling}

To realize a second-moment bound, we need to select singular components with prescribed probabilities while respecting the rank budget on every draw.

For an integer $r\ge0$, the feasible marginal vectors form the polytope
\[
\left\{p\in[0,1]^s:\sum_{i=1}^s p_i\le r\right\}
=\operatorname{conv}\left\{
\mathbf1_S:S\subseteq\{1,\ldots,s\},\ |S|\le r
\right\}.
\]
This is the diagonal form of the convex hull of orthogonal projections of rank at most $r$ \citep{choi1990,overton1992}. Each convex combination on the right is a distribution over admissible subsets, with $p_i$ as the inclusion probability of index $i$. Carath\'eodory's theorem gives a distribution over at most $s+1$ subsets \citep[Section~17]{rockafellar1970}. The systematic sampling construction below produces such a finite distribution directly. It originates in \citet{madow1949} and is also used for singular-component sampling by \citet{barnes2025}.

\begin{lemma}[A subset with prescribed marginals]
\label{lem:subset}
Let $0\le p_i\le1$ for $1\le i\le s$, and suppose $h:=\sum_i p_i\le r$ for an integer $r$. There exists a random subset $S\subseteq\{1,\ldots,s\}$ satisfying
\begin{equation}
\mathbb P(i\in S)=p_i,
\;
|S|\le r~\text{almost surely}.
\label{eq:app-a-subset-marginals}
\end{equation}
Its distribution has at most $s+1$ support points, and $|S|\in\{\lfloor h\rfloor,\lceil h\rceil\}$ almost surely. In particular, if $h$ is an integer, every draw selects exactly $h$ indices.
\end{lemma}

\begin{proof}
Set $t_0=0$ and $t_i=\sum_{j=1}^i p_j$. Draw $U$ uniformly on $[0,1)$ and define $S=\{i:[t_{i-1},t_i)\cap(U+\mathbb Z)\ne\varnothing\}$. Each half-open interval has length at most one and therefore contains at most one point of $U+\mathbb Z$. Its probability of containing a point is its length $p_i$. The intervals are disjoint and cover $[0,h)$, so the number of selected intervals equals the number of lattice points in $[0,h)$. That number is either $\lfloor h\rfloor$ or $\lceil h\rceil$, proving \eqref{eq:app-a-subset-marginals} and the cardinality claims.

As $U$ varies, the subset can change only at the fractional parts of $t_0,\ldots,t_s$. These give at most $s$ nonzero breakpoints, partitioning $[0,1)$ into at most $s+1$ intervals on which the subset is constant. Assign to each subset the total length of the intervals producing it. This gives the required finite distribution. The half-open convention specifies the choices at endpoints, which do not affect the distribution. If $s=0$, take $S=\varnothing$ deterministically.
\end{proof}

\subsection{Feasibility of general quadratic moment bounds}\label{app-toc-a-4}
\label{app:moment-feasibility}

We now allow a rectangular mean and different input and output quadratic forms. The resulting criterion underlies both the fixed-metric result in the main text and the one-step objectives in Appendix~\ref{app:extensions}.

\begin{proposition}[General quadratic moment bounds]
\label{thm:budget}
Let $B\in\mathbb R^{m\times n}$, $P\in\operatorname{Sym}_{++}^m$, $Q\in\operatorname{Sym}_{++}^n$, and let $r$ be an integer with $1\le r\le\min(m,n)$. A random matrix satisfying
\begin{equation}
\mathbb ER=B,
\;
\operatorname{rank}R\le r\ \text{almost surely},
\;
\mathbb E[R^\top PR]\preceq Q
\label{eq:app-a-budget-request}
\end{equation}
exists if and only if
\begin{equation}
B^\top PB\preceq Q,
\;
\operatorname{tr}(Q^{-1}B^\top PB)\le r.
\label{eq:app-a-budget-test}
\end{equation}
Whenever these conditions hold, a distribution supported on at most $\operatorname{rank}(B)+1$ matrices satisfies \eqref{eq:app-a-budget-request}. The moment bound in \eqref{eq:app-a-budget-request} itself guarantees $\mathbb E\|R\|_F^2<\infty$.
\end{proposition}

\begin{proof}
Make the change of variables $D=P^{1/2}BQ^{-1/2}, \; Y=P^{1/2}RQ^{-1/2}$. Both coordinate transformations are invertible, so they preserve rank and finiteness of the second moment. The conditions in \eqref{eq:app-a-budget-request} become
\begin{equation}
\mathbb EY=D,
\;
\operatorname{rank}Y\le r\ \text{almost surely},
\;
\mathbb E[Y^\top Y]\preceq I_n.
\label{eq:app-a-normalized-budget}
\end{equation}

For necessity, the identity $\mathbb E[(Y-D)^\top(Y-D)]=\mathbb E[Y^\top Y]-D^\top D$ gives $D^\top D\preceq I_n$. Lemma~\ref{lem:rank} also gives $\|D\|_F^2\le\mathbb E[\operatorname{rank}Y]\le r$. In the original variables, these are precisely \eqref{eq:app-a-budget-test}.

For sufficiency, take $Y=0$ if $D=0$. Otherwise, write the compact singular value decomposition $D=\sum_{i=1}^s\sigma_i u_iv_i^\top$, where $s=\operatorname{rank}D=\operatorname{rank}B$ and every $\sigma_i>0$. Set $p_i=\sigma_i^2$. The conditions in \eqref{eq:app-a-budget-test} give $p_i\le1$ and $\sum_i p_i\le r$. Choose a random subset $S$ using Lemma~\ref{lem:subset}, and define
\begin{equation}
Y=\sum_{i\in S}\frac{\sigma_i}{p_i}u_iv_i^\top,
\;
R=P^{-1/2}YQ^{1/2}.
\label{eq:app-a-budget-sampler}
\end{equation}
The marginals ensure $\mathbb EY=D$, and orthogonality of the singular vectors gives $\operatorname{rank}Y=|S|\le r$. The same orthogonality removes every cross term in $Y^\top Y$, so $\mathbb E[Y^\top Y] =\sum_{i=1}^s\frac{\sigma_i^2}{p_i}v_iv_i^\top =\sum_{i=1}^s v_iv_i^\top \preceq I_n$. Thus \eqref{eq:app-a-normalized-budget} holds. By Lemma~\ref{lem:subset}, the distribution of $R$ has at most $s+1=\operatorname{rank}(B)+1$ support points and hence finite second moment. Moreover, taking traces in \eqref{eq:app-a-budget-request} gives $\lambda_{\min}(P)\mathbb E\|R\|_F^2\le\operatorname{tr}Q$ for every feasible distribution.
\end{proof}

The criterion controls the budget both in each input direction and in total. After normalization, the first condition bounds every squared singular value by one, while the second bounds their sum by the available rank. The construction in \eqref{eq:app-a-budget-sampler} uses these squared singular values as inclusion probabilities to meet both requirements.

\begin{proof}[Proof of main-text Theorem~\ref{thm:fixed-metric-feasibility} (fixed-metric feasibility)]
Fix $P\succ0$ and set $B_P=P^{1/2}AP^{-1/2}$. For $c>0$, apply Proposition~\ref{thm:budget} with mean $B=A$ and input budget $Q=cP$. This gives
\begin{equation}
\exists R:~ \mathbb ER=A,~
\operatorname{rank}R\le r\ \text{almost surely},~
\mathbb E[R^\top PR]\preceq cP \Longleftrightarrow~
\|B_P\|_2^2\le c,\; \|B_P\|_F^2\le rc.
\label{eq:app-a-fixed-metric-feasibility}
\end{equation}
The support still has at most $\operatorname{rank}(A)+1$ points. If $c=0$, positive definiteness of $P$ forces $R=0$ almost surely and hence $A=0$. Conversely, if $A=0$, the deterministic choice $R=0$ is feasible. The right-hand side of \eqref{eq:app-a-fixed-metric-feasibility} also reduces to $A=0$ when $c=0$, so the equivalence includes this boundary. Consequently, the optimal fixed-metric factor is $c_r(P)=\max\left\{\|B_P\|_2^2,\frac{\|B_P\|_F^2}{r}\right\}$, and a finite-support distribution always attains it.
\end{proof}

\paragraph{Positive semidefinite input budgets.}
If $Q\succeq0$ is allowed to be singular, the complete criterion in Proposition~\ref{thm:budget} becomes
\begin{equation}
B^\top PB\preceq Q,
\;
\operatorname{tr}(Q^\dagger B^\top PB)\le r.
\label{eq:app-a-semidefinite-budget}
\end{equation}
Here $P\succ0$ is still required. The first condition implies the necessary kernel condition $\ker Q\subseteq\ker B$. Every feasible sample must also satisfy $Rx=0$ for all $x\in\ker Q$ almost surely. Whenever the criterion holds, a distribution with at most $\operatorname{rank}(B)+1$ support points remains sufficient.

To prove this extension, first suppose $R$ is feasible. For every $x\in\ker Q$, we have $0\le\mathbb E\|Rx\|_P^2\le x^\top Qx=0$, so $Rx=0$ almost surely. Applying this statement simultaneously to a finite basis of $\ker Q$ shows that the sample vanishes on the whole kernel and that $Bx=0$ there. Jensen's inequality gives $B^\top PB\preceq Q$.

If $Q\ne0$, let the columns of $U\in\mathbb R^{n\times t}$ form an orthonormal basis of $\operatorname{range}(Q)$ and write $Q=U\Lambda U^\top$, where $t=\operatorname{rank}Q$ and $\Lambda\succ0$. The kernel condition gives $R=(RU)U^\top$ and $B=(BU)U^\top$. The original problem is therefore equivalent to the problem with mean $\widetilde B=BU$ and input budget $\Lambda$. Its normalized mean is $\widetilde D=P^{1/2}BU\Lambda^{-1/2}$, and
\[
\|\widetilde D\|_F^2
=\operatorname{tr}(\Lambda^{-1}U^\top B^\top PBU)
=\operatorname{tr}(Q^\dagger B^\top PB).
\]
Lemma~\ref{lem:rank} gives the second necessary condition in \eqref{eq:app-a-semidefinite-budget}. Conversely, if \eqref{eq:app-a-semidefinite-budget} holds, then so does the kernel condition, and $\widetilde D^\top\widetilde D\preceq I_t$ and $\|\widetilde D\|_F^2\le r$. Apply the singular-component construction in \eqref{eq:app-a-budget-sampler} directly to $\widetilde D$, obtaining $\widetilde R$, and set $R=\widetilde R U^\top$. This realizes the original mean, rank constraint, and moment budget. Since $\operatorname{rank}(BU)=\operatorname{rank}B$, the support bound is unchanged. If $Q=0$, condition \eqref{eq:app-a-semidefinite-budget} forces $B=0$, and the only feasible sample is $R=0$ almost surely, so the statement remains valid.

\paragraph{Expected rank and the minimum integer budget.}
The necessity proof uses only $\mathbb E[\operatorname{rank}R]\le r$, whereas the sufficient construction satisfies the samplewise constraint $\operatorname{rank}R\le r$. Thus, for integer $r$, replacing the samplewise bound by an expected-rank bound leaves the criterion in Proposition~\ref{thm:budget} unchanged. If $B\ne0$, $Q\succ0$, and $B^\top PB\preceq Q$, the least positive integer rank budget is $\lceil\operatorname{tr}(Q^{-1}B^\top PB)\rceil$. The same conclusion holds for a positive semidefinite budget satisfying \eqref{eq:app-a-semidefinite-budget}, with $Q^{-1}$ replaced by $Q^\dagger$.

\subsection{The optimal growth factor and the stability threshold}\label{app-toc-a-5}
\label{app:optimal-factor}

We now optimize the metric. The key is to construct one family of coordinate transformations that simultaneously approaches the spectral lower bounds for the operator and Frobenius norms.

\begin{proof}[Proof of main-text Theorem~\ref{thm:optimal-mean-square-growth} (optimal growth factor)]
If $A=0$, the deterministic choice $R=0$ proves all claims. Assume henceforth that $A\ne0$. For fixed $P\succ0$, write $B_P=P^{1/2}AP^{-1/2}$ and define $H_\mu(P)=\lambda_{\max}\!\left( P^{-1/2}\mathcal M_\mu^*(P)P^{-1/2}\right)$. By \eqref{eq:app-a-fixed-metric-feasibility}, for any $c>0$ an admissible distribution with $H_\mu(P)\le c$ exists if and only if $\|B_P\|_2^2\le c$ and $\|B_P\|_F^2\le rc$. Thus $\inf_\mu H_\mu(P)=c_r(P)>0$, and a finite-support distribution attains this value. Applying Lemma~\ref{lem:lyapunov} and exchanging the two infima gives
\begin{equation}
q_r(A)
=\inf_{\mu\in\mathcal C_r(A)}\inf_{P\succ0}H_\mu(P) =\inf_{P\succ0}\inf_{\mu\in\mathcal C_r(A)}H_\mu(P) =\inf_{P\succ0}
\max\left\{\|B_P\|_2^2,\frac{\|B_P\|_F^2}{r}\right\}.
\label{eq:app-a-metric-reduction}
\end{equation}
This also proves the metric reduction at the end of Section~3 in the main text. Every matrix $B$ similar to $A$ satisfies $\|B\|_2^2\ge\rho(A)^2, \; \|B\|_F^2\ge\mathcal E(A)$. The first inequality is the spectral-radius bound. For the second, a complex Schur decomposition transforms $B$ by unitary similarity into an upper triangular matrix whose diagonal contains the eigenvalues of $A$. Deleting its strictly upper triangular part leaves squared Frobenius norm $\mathcal E(A)$. Equation~\eqref{eq:app-a-metric-reduction} therefore gives $q_r(A)\ge\max\{\rho(A)^2,\mathcal E(A)/r\}$.

To approach both bounds simultaneously, we may use arbitrary real invertible coordinate transformations. Indeed, the polar decomposition $S=UP^{1/2}$, with $P=S^\top S\succ0$ and $U$ orthogonal, gives $SAS^{-1}=U(P^{1/2}AP^{-1/2})U^\top$. The matrices on the two sides have the same operator and Frobenius norms.

Choose a real Jordan form $A=VJV^{-1}$. Its diagonal spectral blocks are real scalars or two-dimensional blocks $C(a,b)=\left(\begin{smallmatrix}a&-b\\b&a\end{smallmatrix}\right)$ corresponding to conjugate eigenvalues $a\pm ib$. Each such block is normal. For a Jordan chain containing $\ell$ spectral blocks, scale the corresponding coordinates by $\varepsilon^{\ell-1},\ldots,\varepsilon,1$, using the same factor for both coordinates of each two-dimensional block. Let $D_\varepsilon$ combine these scalings over all chains, and set $S_\varepsilon=D_\varepsilon V^{-1}$. Every Jordan link is then multiplied by $\varepsilon$, so $S_\varepsilon AS_\varepsilon^{-1}\longrightarrow N, \|N\|_2^2=\rho(A)^2, \; \|N\|_F^2=\mathcal E(A)$, where $N$ is the normal block-diagonal matrix obtained by deleting every Jordan link. Norm continuity and \eqref{eq:app-a-metric-reduction} yield $q_r(A)=\max\left\{\rho(A)^2,\frac{\mathcal E(A)}r\right\}$.

Given $c>q_r(A)$, choose $\varepsilon>0$ small enough that $\|S_\varepsilon AS_\varepsilon^{-1}\|_2^2<c$ and $\|S_\varepsilon AS_\varepsilon^{-1}\|_F^2/r<c$. Set $P=S_\varepsilon^\top S_\varepsilon$. Polar decomposition and Proposition~\ref{thm:budget} give an admissible distribution with at most $\operatorname{rank}(A)+1$ support points satisfying $\mathcal M_\mu^*(P)\preceq cP$. Lemma~\ref{lem:lyapunov} then gives the finite-step bound in the main text, $\mathbb E\|\Phi_kx\|_2^2\le\kappa_2(P)c^k\|x\|_2^2$.

If $A$ is diagonalizable over $\mathbb C$, its real Jordan form has no links, so a fixed real similarity gives $N$ itself. Since $A\ne0$ then implies $q_r(A)>0$, the same moment construction applies directly at $c=q_r(A)$ and produces an optimal distribution. The case $A=0$ was handled at the start.
\end{proof}

\begin{corollary}[The exact rank threshold for stability]
\label{cor:stability}
There exists a mean-square exponentially stable distribution in $\mathcal C_r(A)$ if and only if
\begin{equation}
\rho(A)<1,
\;
\mathcal E(A)<r.
\label{eq:app-a-stability-test}
\end{equation}
For a Schur stable matrix, meaning $\rho(A)<1$, the least positive integer rank budget permitting stable sampling is
\begin{equation}
r_{\min}(A)=\lfloor\mathcal E(A)\rfloor+1.
\label{eq:app-a-minimum-stable-rank}
\end{equation}
Whenever a stable distribution exists, a finite-support stable distribution exists.
\end{corollary}

\begin{proof}
By Lemma~\ref{lem:lyapunov}, a stable distribution satisfies $\rho(\mathcal M_\mu)<1$. Since $q_r(A)\le\rho(\mathcal M_\mu)$, main-text Theorem~\ref{thm:optimal-mean-square-growth} implies \eqref{eq:app-a-stability-test}. Conversely, \eqref{eq:app-a-stability-test} gives $q_r(A)<1$. Choosing $q_r(A)<c<1$, the finite-support construction and quadratic bound in main-text Theorem~\ref{thm:optimal-mean-square-growth} yield a stable distribution. The least positive integer strictly greater than $\mathcal E(A)$ is \eqref{eq:app-a-minimum-stable-rank}. It does not exceed $d$, since Schur stability gives $\mathcal E(A)<d$.
\end{proof}

If rank zero is allowed, its only feasible mean is $A=0$. At the other end, if $r\ge\operatorname{rank}A$, the deterministic distribution $R=A$ attains the optimum. Indeed, $A$ has at most $\operatorname{rank}A$ nonzero eigenvalues counted according to algebraic multiplicity. Hence $\mathcal E(A)\le\operatorname{rank}(A)\rho(A)^2$ and $q_r(A)=\rho(A)^2$.

\section{Attainment and construction of optimal distributions}\label{app-toc-b}
\label{app:attainment}

This appendix proves the attainment theorem in the main text. The argument examines how the rank budget can be allocated once the growth factor reaches its optimal value. We first put the samples into a common upper block triangular form and analyze equality in the rank bound. This gives a restriction on the length of the zero Jordan chains. We then construct finite-support distributions satisfying that restriction and examine optimal positive definite quadratic bounds and uniform finite-step estimates at the optimal rate.

Following the main text, write
\[
q=q_r(A)=\max\left\{\rho(A)^2,\frac{\mathcal E(A)}r\right\},
\;
\mathcal E(A)=\sum_{i=1}^d|\lambda_i(A)|^2.
\]
Let $\nu_0(A)$ denote the largest Jordan block size at zero, with $\nu_0(A)=0$ when $A$ is invertible. Let $b(A)$ count the nonzero real eigenvalues and nonreal conjugate pairs with algebraic multiplicity, counting each pair once. Thus $b(A)$ counts the nonzero one- or two-dimensional diagonal spectral blocks in a real Jordan form. A nonzero real Jordan chain of length $\ell$ contributes $\ell$ to this count, as does a conjugate pair of nonreal Jordan chains of length $\ell$.

For a random matrix $R$ with finite second moment, let $\mathcal M_R(X)=\mathbb E[RXR^\top]$ be its second-moment operator and $\mathcal M_R^*(P)=\mathbb E[R^\top PR]$ its adjoint. Both act on real symmetric matrices and have the same spectral radius. Lemma~\ref{lem:lyapunov} establishes $g(\mu)=\rho(\mathcal M_R)$.

A fixed real similarity preserves samplewise rank, finite second moments, and the asymptotic growth factor. Specifically, for $\Gamma_S(X)=SXS^\top$, $\mathcal M_{SRS^{-1}} =\Gamma_S\mathcal M_R\Gamma_S^{-1}$. We may therefore choose coordinates to suit each argument and transform the resulting construction back afterward.

\subsection{Common block structure and positive definite eigenmatrices}\label{app-toc-b-1}
\label{app:block-structure}

When all samples preserve a subspace, their actions on that subspace and its quotient can be studied separately. The next two lemmas explain how a common triangular decomposition determines the growth factor and why an irreducible diagonal family admits a positive definite eigenmatrix. Together, they allow us to analyze equality in the rank bound on each diagonal block.

\begin{lemma}[Mean-square growth of upper block triangular matrices]
\label{lem:block-radius}
Let $R$ be a random real square matrix with finite second moment. Suppose $R$ is upper block triangular almost surely in a fixed decomposition, with diagonal blocks $D_1,\ldots,D_s$. Then $\rho(\mathcal M_R) =\max_{1\le j\le s}\rho(\mathcal M_{D_j})$. The diagonal blocks within a sample may be dependent.
\end{lemma}

\begin{proof}
Take independent copies of the full matrix $R$. Let $G_k$ be their product over the first $k$ steps and $H_{j,k}$ its $j$th diagonal block. The latter is the product of the corresponding diagonal block samples. Embed a unit vector $z$ in block $j$ of the full space. The $j$th block of its image under $G_k$ is $H_{j,k}z$, so $\sup_{\|z\|_2=1}\mathbb E\|H_{j,k}z\|_2^2 \le \sup_{\|x\|_2=1}\mathbb E\|G_kx\|_2^2$. Taking $k$th roots and letting $k\to\infty$, the growth characterization in Lemma~\ref{lem:lyapunov} gives $\rho(\mathcal M_{D_j})\le\rho(\mathcal M_R)$. This proves the lower bound.

For the upper bound, fix $c>\max_j\rho(\mathcal M_{D_j})$ with $c>0$. The quadratic characterization in Lemma~\ref{lem:lyapunov} gives, for each $j$, a matrix $P_j\succ0$ such that $\mathbb E[D_j^\top P_jD_j]\preceq cP_j$. Apply the block diagonal similarity with blocks $P_j^{1/2}$. Retaining the notation $D_j$ for the transformed diagonal blocks, we now have $\mathbb E[D_j^\top D_j]\preceq cI$. Apply the further similarity $S_\varepsilon =\operatorname{diag}\bigl( \varepsilon^{s-1}I,\varepsilon^{s-2}I,\ldots,I \bigr)$. The strictly upper block $(i,j)$ is multiplied by $\varepsilon^{j-i}$. As $\varepsilon\downarrow0$, the transformed random matrix converges in mean squared Frobenius norm to its block diagonal part. By Cauchy--Schwarz, its expected right Gram matrix also converges, with limit $\operatorname{diag}\bigl( \mathbb E[D_1^\top D_1],\ldots, \mathbb E[D_s^\top D_s] \bigr)\preceq cI$. Each similarity with $\varepsilon>0$ preserves the growth factor, which is bounded above by the operator norm of the corresponding one-step right second moment. Letting $\varepsilon\downarrow0$ gives $\rho(\mathcal M_R)\le c$. Finally, decrease $c$ to $\max_j\rho(\mathcal M_{D_j})$.
\end{proof}

\begin{lemma}[Positive definite eigenmatrix]
\label{lem:perron}
Every positive linear map $T:\operatorname{Sym}^m\to\operatorname{Sym}^m$ has a nonzero $P\succeq0$ satisfying $T(P)=\rho(T)P$. If, in addition, $T(X)=\mathbb E[D^\top XD]$, $D$ has finite second moment, and the topological support of its distribution has no proper nonzero common invariant real subspace, then $P$ can be chosen positive definite.
\end{lemma}

\begin{proof}
For $\varepsilon>0$, set $T_\varepsilon(X)=T(X)+\varepsilon\operatorname{tr}(X)I_m$. The continuous map
\[
X\longmapsto
\frac{T_\varepsilon(X)}{\operatorname{tr}(T_\varepsilon(X))}
\]
maps the compact convex set $\{X\succeq0:\operatorname{tr}X=1\}$ into itself. Brouwer's fixed point theorem \citep[Section~3.2]{deimling1985} gives a fixed point $P_\varepsilon\succ0$ with $\operatorname{tr}P_\varepsilon=1$ and $T_\varepsilon(P_\varepsilon)=\lambda_\varepsilon P_\varepsilon$ for some $\lambda_\varepsilon>0$.

Since $P_\varepsilon\succ0$, there is a constant $a_\varepsilon$ such that every symmetric $X$ with $\|X\|_2\le1$ satisfies $-a_\varepsilon P_\varepsilon\preceq X \preceq a_\varepsilon P_\varepsilon$. Positivity of $T_\varepsilon$ and iteration give $\|T_\varepsilon^k\| \le a_\varepsilon\|P_\varepsilon\|_2\lambda_\varepsilon^k$. The spectral radius formula implies $\rho(T_\varepsilon)\le\lambda_\varepsilon$. Equality follows because $\lambda_\varepsilon$ is an eigenvalue. Compactness of the trace-one positive semidefinite matrices gives a subsequence along which $P_\varepsilon\to P\succeq0$ as $\varepsilon\downarrow0$, with $\operatorname{tr}P=1$. Continuity of the spectral radius for finite-dimensional linear maps then yields $T(P)=\rho(T)P$.

Now assume the stated expectation representation of $T$. For every $x\in\ker P$, $\mathbb E\|P^{1/2}Dx\|_2^2 =x^\top T(P)x=0$. Applying this identity to a basis of $\ker P$ shows that the kernel is invariant under $D$ almost surely. Continuity extends this invariance to the entire topological support. Since $P\ne0$, its kernel is not the whole space, and irreducibility excludes a proper nonzero invariant subspace. Hence $\ker P=\{0\}$ and $P\succ0$.
\end{proof}

\subsection{Rank equality and the length of zero Jordan chains}\label{app-toc-b-2}
\label{app:rank-equality}

We restate the attainment classification and first prove the necessity of the spectral condition in its third case.

\begin{theorem}[Attainment of the optimum; main-text Theorem~\ref{thm:attainment}]
\label{thm:attainment-appendix}
For every $A\in\mathbb R^{d\times d}$ and integer $1\le r\le d$, attainment of $q_r(A)$ is characterized as follows.
\begin{enumerate}
\setlength{\itemsep}{0pt}
\item If $\rho(A)=0$, the optimum is attained.
\item If $\rho(A)^2>\mathcal E(A)/r$, the optimum is attained.
\item If $\mathcal E(A)/r\ge\rho(A)^2>0$, the optimum is attained if and only if $\nu_0(A)\le b(A)+1$. \end{enumerate}
In every attainable case, a finite-support optimal distribution exists.
\end{theorem}

\begin{proof}[Necessity in the third case]
Suppose $q=\mathcal E(A)/r\ge\rho(A)^2>0$ and an admissible distribution attains $g(\mu)=q$. We identify its common block decomposition, the equality conditions in the rank bound, and the resulting zero layers.

Choose a common invariant subspace of the support having the least positive dimension, and repeat in the quotient. This gives a common invariant flag whose successive quotient families are irreducible over $\mathbb R$. An adapted orthonormal basis makes all samples upper block triangular. Denote the irreducible diagonal blocks by $D_j$ and their means by $B_j=\mathbb ED_j$.

Lemmas~\ref{lem:block-radius} and~\ref{lem:perron} give numbers $q_j\le q$ and matrices $P_j\succ0$ such that $\mathbb E[D_j^\top P_jD_j]=q_jP_j$. Normalize the random blocks and their means by setting
\begin{equation}
Y_j=q^{-1/2}P_j^{1/2}D_jP_j^{-1/2}, F_j=\mathbb EY_j
=q^{-1/2}P_j^{1/2}B_jP_j^{-1/2}.
\label{eq:app-b-05}
\end{equation}
Then $C_j:=\mathbb E[Y_j^\top Y_j]=(q_j/q)I\preceq I$. The rank inequality in Lemma~\ref{lem:rank} and a Schur decomposition imply $\mathbb E\operatorname{rank}D_j \ge\|F_j\|_F^2 \ge\frac{\mathcal E(B_j)}q$. The rank of an upper block triangular matrix is at least the sum of the ranks of its diagonal blocks. Moreover, the spectra of the $B_j$ together form the spectrum of $A$, including algebraic multiplicities. Consequently,
\begin{equation}
r
\ge\mathbb E\operatorname{rank}R
\ge\sum_j\mathbb E\operatorname{rank}D_j
\ge\sum_j\|F_j\|_F^2
\ge\frac{\mathcal E(A)}q=r.
\label{eq:app-b-06}
\end{equation}
Every inequality is an equality. In particular, the squared Frobenius norm of each $F_j$ equals the sum of its squared eigenvalue moduli. Its complex Schur form therefore has no strictly upper triangular entries, so $F_j$ is normal.

We next show that each $B_j$ is either zero or invertible and that $B_j=0$ forces $D_j=0$ almost surely. If all eigenvalues of $B_j$ vanish, equality in \eqref{eq:app-b-06} gives $\mathbb E\operatorname{rank}D_j=0$, hence $D_j=0$ almost surely. Now suppose $B_j$ has a nonzero eigenvalue but $F_j$ is singular. Let $\Pi_j$ be the orthogonal projector onto $\operatorname{range}Y_j$. Expanding a square and using $\Pi_jY_j=Y_j$ yields
\begin{equation}
\mathbb E\operatorname{rank}Y_j-\|F_j\|_F^2
={}\mathbb E\|\Pi_j-Y_jF_j^\top\|_F^2 +\operatorname{tr}\bigl(F_j(I-C_j)F_j^\top\bigr).
\label{eq:app-b-07}
\end{equation}
Indeed, the three terms in the squared norm expand as
\[
\mathbb E\|\Pi_j\|_F^2
=\mathbb E\operatorname{rank}Y_j,~~
-2\mathbb E\operatorname{tr}(\Pi_jY_jF_j^\top)
=-2\|F_j\|_F^2,~~
\mathbb E\|Y_jF_j^\top\|_F^2
=\operatorname{tr}(F_jC_jF_j^\top).
\]
By \eqref{eq:app-b-06}, the left-hand side of \eqref{eq:app-b-07} is zero, and both terms on its right-hand side are nonnegative. Hence $\Pi_j=Y_jF_j^\top$ almost surely. For $x\in\ker F_j^\top$, this gives $\Pi_jx=0$, and therefore $\operatorname{range}Y_j \subseteq(\ker F_j^\top)^\perp =\operatorname{range}F_j ~\text{almost surely}$. Because $F_j$ is nonzero and singular, $\operatorname{range}F_j$ is a proper nonzero subspace. The inclusion makes it invariant under every $Y_j$ almost surely and, by continuity, under the entire support. This contradicts irreducibility, which is preserved by the similarity in \eqref{eq:app-b-05}. The singular case is therefore impossible.

Let $k$ be the number of invertible mean blocks. Merge each consecutive group of zero-mean diagonal blocks into one block. Although its individual diagonal blocks already vanish, an aggregated block may initially contain random strictly upper triangular entries. Applying the block rank inequality to the coarser decomposition gives
\[
\mathbb E\operatorname{rank}R
\ge
\sum_{j:B_j\text{ invertible}}\mathbb E\operatorname{rank}D_j
+\sum_{\text{aggregated blocks }Z}\mathbb E\operatorname{rank}Z.
\]
The first sum already equals $r$ by \eqref{eq:app-b-06}. Thus every aggregated block satisfies $\mathbb E\operatorname{rank}Z=0$ and vanishes almost surely. Taking expectations, $A$ has an upper block triangular representation with at most $k+1$ zero diagonal blocks interleaved with the $k$ invertible blocks: $0,\ B_1,\ 0,\ B_2,\ \ldots,\ B_k,\ 0$. Zero-dimensional layers at the ends or between the invertible blocks may be omitted. The remaining off-diagonal blocks are unrestricted.

To translate this structure into a bound on the zero Jordan chains, consider the resolvent $(zI-A)^{-1}$ near $z=0$. Expanding its $(i,j)$ block along strictly increasing block paths, each term has the form
\[
(zI-A_{i_0i_0})^{-1}A_{i_0i_1}
(zI-A_{i_1i_1})^{-1}\cdots \cdot
A_{i_{\ell-1}i_\ell}(zI-A_{i_\ell i_\ell})^{-1},
\]
where $i=i_0<i_1<\cdots<i_\ell=j$; a diagonal block corresponds to a path with no off-diagonal factor. The resolvent of an invertible diagonal block is analytic near zero, whereas each zero diagonal block contributes $z^{-1}I$. An increasing path visits each block at most once and therefore contains at most $k+1$ factors $z^{-1}$. The full resolvent consequently has a pole of order at most $k+1$ at zero.

For a zero Jordan block, direct expansion gives $(zI-J_m(0))^{-1} =\sum_{\ell=0}^{m-1}z^{-\ell-1}J_m(0)^\ell$, whose pole order is exactly $m$. Similarity preserves pole order, so the pole order of $(zI-A)^{-1}$ at zero is $\nu_0(A)$, with order zero when $A$ is invertible. Hence $\nu_0(A)\le k+1\le b(A)+1$. The last inequality holds because each invertible real mean block contains at least one nonzero real eigenvalue or one nonreal conjugate pair, and these counts add over diagonal blocks. This proves necessity in the third case.
\end{proof}

\subsection{Finite-support constructions in all three regimes}\label{app-toc-b-3}
\label{app:attainment-constructions}

We now complete the sufficiency proof of Theorem~\ref{thm:attainment-appendix}. We first handle the case governed by the chain-length criterion, followed by nilpotent means and the regime in which the spectral-radius term is strictly larger.

\begin{proof}[Sufficiency in the third case]
Suppose $q=\mathcal E(A)/r\ge\rho(A)^2>0$ and $\nu_0(A)\le b(A)+1$, and set $b=b(A)$. Work in real Jordan coordinates. Let $T\in\mathbb R^{n\times n}$ denote the invertible part of the mean. Its diagonal consists of $b$ normal spectral blocks $C_1,\ldots,C_b$, where block $j$ has dimension $h_j\in\{1,2\}$ and satisfies $C_j^\top C_j=\omega_j^2I_{h_j}$ with $\omega_j>0$.

The strictly upper blocks of $T$ retain the Jordan links at nonzero eigenvalues. Assign every coordinate of block $j$ the marginal selection probability $p_j=\omega_j^2/q$. Since $q\ge\rho(A)^2$ and $rq=\mathcal E(A)$, $0<p_j\le1, \; \sum_{j=1}^b h_jp_j =\frac{\mathcal E(A)}q=r$. Lemma~\ref{lem:subset} supplies a finite-support random subset $S$ with these marginals and $|S|=r$ almost surely.

To incorporate the zero Jordan chains, place their coordinates in spaces $H_0,\ldots,H_b$ so that the nilpotent part contains only links $N_j:H_j\to H_{j-1}$ for $1\le j\le b$. Specifically, place the first, second, and subsequent coordinates of each chain in $H_0,H_1,\ldots$. The condition $\nu_0(A)\le b+1$ ensures this is possible, and empty layers are allowed. Order the full coordinates as
\begin{equation}
H_0,\ C_1,\ H_1,\ C_2,\ \ldots,\ C_b,\ H_b.
\label{eq:app-b-11}
\end{equation}

For coordinate $\ell$ of spectral block $j$, let $e_{j\ell}$ be the corresponding standard basis vector in the full space. Let $t_{j\ell}^\top$ be the corresponding row of $T$, extended by zeros on all $H_i$. Independently of $S$, choose finite-valued random vectors $\xi_{j\ell}\in H_j$ satisfying $\mathbb E\xi_{j\ell}=0$ and $\mathbb E[\xi_{j\ell}\xi_{j\ell}^\top]=I_{H_j}$. Independent Rademacher coordinates suffice; use the zero vector when $H_j$ is empty. In the following formula, embed $\xi_{j\ell}$ in $H_j$ and $N_j\xi_{j\ell}$ in $H_{j-1}$, and define
\begin{equation}
F_{j\ell}
=\left(e_{j\ell}+\frac{N_j\xi_{j\ell}}{h_jp_j}\right)
  \left(\frac{t_{j\ell}}{p_j}+\xi_{j\ell}\right)^\top, R=\sum_{(j,\ell)\in S}F_{j\ell}.
\label{eq:app-b-12}
\end{equation}
We establish its rank, mean, and growth factor in turn.

First, each $F_{j\ell}$ is an outer product and has rank at most one. Exactly $r$ coordinates are selected on every draw, so $\operatorname{rank}R\le r$ almost surely.

Second, expand the outer product in \eqref{eq:app-b-12}. The two cross terms containing a single factor $\xi_{j\ell}$ have zero mean. Independence of the subset and random vectors gives
\[
\mathbb E\bigl[\mathbf 1_{\{(j,\ell)\in S\}}F_{j\ell}\bigr]
=p_j\left(
\frac{e_{j\ell}t_{j\ell}^\top}{p_j}
+\frac{N_j}{h_jp_j}
\right) =e_{j\ell}t_{j\ell}^\top+\frac1{h_j}N_j.
\]
Here $N_j$ occupies the block $(H_{j-1},H_j)$. Summing over the $h_j$ coordinates of block $j$ gives one full copy of $N_j$. Summing over all $j$ then supplies every row of $T$ and all zero-chain links. Thus $\mathbb ER$ is the prescribed mean.

Finally, every sample is upper block triangular in the order \eqref{eq:app-b-11}. Its diagonal blocks on the zero layers $H_i$ vanish, while its diagonal block on $C_j$ is
\[
D_j=p_j^{-1}\Pi_jC_j,
\;
\Pi_j=\operatorname{diag}
\bigl(\mathbf 1_{\{(j,\ell)\in S\}}\bigr)_{\ell=1}^{h_j}.
\]
Using $\Pi_j^2=\Pi_j$ and $\mathbb E\Pi_j=p_jI_{h_j}$, $\mathbb E[D_j^\top D_j] =p_j^{-1}C_j^\top C_j =qI_{h_j}$. Each diagonal block therefore has growth factor at most $q$, and Lemma~\ref{lem:block-radius} gives $\rho(\mathcal M_R)\le q$. The optimal-value lower bound proved in Section~\ref{app:optimal-factor} gives $\rho(\mathcal M_R)\ge q_r(A)=q$, so this distribution attains the optimum. Both the subset and random vectors have finitely many possible values, giving finite support and finite second moment. The inverse similarity yields the construction in the original coordinates.
\end{proof}

\begin{proof}[The first two cases]
If $\rho(A)=0$, then $A$ is nilpotent and a real similarity puts it in strictly upper triangular form. For $A\ne0$, assign each nonzero entry $A_{ij}$ a positive probability $p_{ij}$, with all probabilities summing to one, and take $R=(A_{ij}/p_{ij})e_ie_j^\top$ with probability $p_{ij}$. This finite distribution has mean $A$, and every sample has rank one. All samples are strictly upper triangular, so every product of $d$ samples vanishes. Hence $\rho(\mathcal M_R)=0=q_r(A)$. For $A=0$, take $R=0$. Since $r\ge1$, these constructions satisfy every admissible rank budget.

Now suppose $q=\rho(A)^2>\mathcal E(A)/r$. Choose real coordinates in which $A=\operatorname{diag}(A_{\mathrm p},A_{\mathrm s})$, where $A_{\mathrm p}$ contains the full generalized eigenspaces associated with eigenvalues of modulus $\rho(A)$ and $A_{\mathrm s}$ contains the remaining spectrum. Let $m=\dim A_{\mathrm p}$. The inequality $\mathcal E(A)=m\rho(A)^2+\mathcal E(A_{\mathrm s}) <r\rho(A)^2$ gives $m<r$, $\rho(A_{\mathrm s})^2<q$, and $\mathcal E(A_{\mathrm s})/(r-m)<q$. The optimal-value formula and finite-support construction for every strictly larger factor in Section~\ref{app:optimal-factor} supply a finite distribution $R_{\mathrm s}$ with mean $A_{\mathrm s}$, rank at most $r-m$, and growth factor strictly below $q$. Set $R=\operatorname{diag}(A_{\mathrm p},R_{\mathrm s})$. Its mean is $A$, and every sample has rank at most $r$. The deterministic block $A_{\mathrm p}$ has growth factor $\rho(A_{\mathrm p})^2=q$, including when it has nontrivial Jordan chains. Lemma~\ref{lem:block-radius} therefore gives growth factor $q$ for the full distribution. Together with the necessity proof in Section~\ref{app:rank-equality}, this completes Theorem~\ref{thm:attainment-appendix}.
\end{proof}

\subsection{Optimal positive definite quadratic bounds}\label{app-toc-b-4}
\label{app:quadratic-attainment}

An optimal distribution attains the asymptotic growth factor. Requiring a fixed positive definite metric that controls the expected squared length at every step with this same factor imposes stronger spectral conditions. The next proposition gives their exact form.

\begin{proposition}[Attainment by a positive definite quadratic bound]
\label{prop:quadratic-attainment}
Let $q=q_r(A)$. There exist an admissible distribution $\mu$ and $P\succ0$ satisfying
\begin{equation}
\mathcal M_\mu^*(P)=\mathbb E[R^\top PR]\preceq qP
\label{eq:app-b-16}
\end{equation}
if and only if the corresponding condition below holds:
\begin{enumerate}
\setlength{\itemsep}{0pt}
\item If $\mathcal E(A)/r\ge\rho(A)^2$, then $A$ is diagonalizable over $\mathbb C$.
\item If $\rho(A)^2>\mathcal E(A)/r$, then all eigenvalues of modulus $\rho(A)$ are semisimple, meaning their Jordan blocks have size one.
\end{enumerate}
Whenever \eqref{eq:app-b-16} is feasible, a finite-support distribution suffices.
\end{proposition}

\begin{proof}
First suppose $q=0$. Since $P\succ0$, condition \eqref{eq:app-b-16} forces $R=0$ almost surely and hence $A=0$. Conversely, $A=0$ admits the zero distribution with any $P\succ0$. A nilpotent matrix is diagonalizable over $\mathbb C$ exactly when it is zero, which proves the statement for $q=0$.

Now let $q>0$. By Proposition~\ref{thm:budget} with $B=A$ and $Q=qP$, a given $P\succ0$ admits a distribution satisfying \eqref{eq:app-b-16} if and only if
\begin{equation}
\|P^{1/2}AP^{-1/2}\|_2^2\le q, \|P^{1/2}AP^{-1/2}\|_F^2\le rq.
\label{eq:app-b-17}
\end{equation}
In the first regime, $rq=\mathcal E(A)$, so equality must hold between the squared Frobenius norm and the sum of squared eigenvalue moduli. The transformed matrix is therefore normal, and $A$ is diagonalizable over $\mathbb C$.

Conversely, a real matrix diagonalizable over $\mathbb C$ is real-similar to a normal block diagonal matrix with scalar real blocks and blocks $\left(\begin{smallmatrix}a&-b\\b&a\end{smallmatrix}\right)$. Its squared operator and Frobenius norms are $\rho(A)^2$ and $\mathcal E(A)$, respectively. Let $S$ be the similarity matrix. Its polar decomposition is $S=UP^{1/2}$ with $P=S^\top S\succ0$, so $P^{1/2}AP^{-1/2}$ is orthogonally similar to $SAS^{-1}$ and satisfies \eqref{eq:app-b-17}.

In the second regime, $q=\rho(A)^2>0$. The first inequality in \eqref{eq:app-b-17} makes all powers of $P^{1/2}AP^{-1/2}/\rho(A)$ bounded. A nontrivial Jordan block at an eigenvalue on the unit circle would yield unbounded polynomial growth. Thus all eigenvalues of modulus $\rho(A)$ must be semisimple.

Conversely, if these eigenvalues are semisimple, put their spectral part into one real normal block and the remaining Jordan blocks into another. Scale only the Jordan links in the latter block. All of its eigenvalues have modulus strictly below $\rho(A)$, so its operator norm is also below $\rho(A)$ for sufficiently small positive scaling. At the same time, the full squared Frobenius norm approaches $\mathcal E(A)<r\rho(A)^2$. Some positive scaling therefore gives a real similarity satisfying both inequalities in \eqref{eq:app-b-17}. Polar decomposition again supplies the required $P\succ0$.

In both regimes, the fixed-metric construction from Proposition~\ref{thm:budget} provides a finite-support distribution.
\end{proof}

\subsection{Rank-one structure and a uniform bound with one nonzero eigenvalue}\label{app-toc-b-5}
\label{app:rank-one-rigidity}

When the mean has exactly one nonzero eigenvalue of algebraic multiplicity one, equality in the rank bound also determines the sample structure of every optimal distribution. This gives a uniform geometric estimate at the optimal rate, including cases without an optimal positive definite quadratic bound.

\begin{corollary}[One nonzero eigenvalue]
\label{cor:rigid-rank-one}
Suppose $A\in\mathbb R^{d\times d}$ has exactly one nonzero eigenvalue $a\ne0$, with algebraic multiplicity one. Then $a\in\mathbb R$, $q_1(A)=a^2$, and an optimal rank-one distribution exists if and only if
\begin{equation}
A^2(A-aI_d)=0.
\label{eq:app-b-18}
\end{equation}
Equivalently, every zero Jordan block has size at most two. A finite-support distribution suffices whenever attainment is possible. Moreover, every attaining distribution $\mu$ has a finite constant $C_\mu$ such that its independent samples satisfy
\begin{equation}
\mathbb E\|R_{k-1}\cdots R_0x\|_2^2
\le C_\mu|a|^{2k}\|x\|_2^2,
\; k\ge0.
\label{eq:app-b-19}
\end{equation}
\end{corollary}

\begin{proof}
Nonreal eigenvalues of a real matrix occur in conjugate pairs, so the unique nonzero eigenvalue $a$ is real. The optimal-value formula gives $q_1(A)=a^2$. Since $b(A)=1$, Theorem~\ref{thm:attainment-appendix} makes attainment equivalent to $\nu_0(A)\le2$. Applying the polynomial $z^2(z-a)$ to the Jordan blocks proves its equivalence to \eqref{eq:app-b-18}.

Fix any optimal distribution. Its equality decomposition in \eqref{eq:app-b-06} has exactly one invertible mean block, the scalar $a$. Its scalar sample $D$ satisfies $\mathbb ED=a$ and $\mathbb ED^2\le a^2$, so $D=a$ almost surely. The aggregated zero blocks before and after it vanish. The rank-one constraint and nonzero pivot $a$ then force every sample to have the form
\[
R=
\begin{pmatrix}
0&u&uv^\top/a\\
0&a&v^\top\\
0&0&0
\end{pmatrix}
~\text{almost surely}
\]
in a fixed basis, with either outer block allowed to be absent. Direct multiplication shows that any three matrices $R,S,T$ of this form satisfy $RST=aRT$. This identity is preserved by a common similarity. Hence, in the original coordinates, $R_{k-1}\cdots R_0 =a^{k-2}R_{k-1}R_0 ~\text{almost surely},\; k\ge2$. Let $m_\mu=\mathbb E\|R\|_2^2<\infty$. For $k\ge2$, the endpoint samples are independent, giving $\mathbb E\|R_{k-1}\cdots R_0\|_2^2 \le |a|^{2k-4}m_\mu^2$. Including $k=0,1$, inequality \eqref{eq:app-b-19} follows with $C_\mu =\max\left\{ 1,\frac{m_\mu}{|a|^2},\frac{m_\mu^2}{|a|^4} \right\}$.

A finite-support distribution can also be constructed directly. Under \eqref{eq:app-b-18}, a real similarity gives $A=aee^\top+N$ with $\|e\|_2=1$, $N^2=0$, and $Ne=N^\top e=0$. The spaces $L=\operatorname{range}N$ and $H=\operatorname{range}N^\top$ are orthogonal to each other and to $e$. In particular, $L\perp H$ follows from $N^2=0$, since $\langle Nx,N^\top y\rangle=y^\top N^2x=0$ for all $x,y$.

If $N=0$, take $R=A$. Otherwise, let $h_1,\ldots,h_s$ be an orthonormal basis of $H$, and let $z$ take the $2s$ values $\pm\sqrt{s}\,h_j$ with equal probabilities. Then $\mathbb Ez=0$ and $\mathbb E[zz^\top]=\Pi_H$. Define $R=(ae+Nz)(e+z)^\top$. Every sample has rank one, and $\mathbb ER=aee^\top+N\Pi_H=A$. For any two support vectors $z,w$, the orthogonality relations give $(e+z)^\top(ae+Nw)=a$. Thus $R_{k-1}\cdots R_0 =a^{k-1}(ae+Nz_{k-1})(e+z_0)^\top, \; k\ge1$. Finite support bounds both endpoint factors, so the growth factor is at most $a^2$; the optimal-value lower bound gives equality. Transforming back yields a finite-support optimal distribution for the original mean.
\end{proof}

For example, if $A=\operatorname{diag}(a,J_2(0))$ with $a\ne0$, every optimal rank-one distribution satisfies \eqref{eq:app-b-19}, whereas Proposition~\ref{prop:quadratic-attainment} excludes an optimal positive definite quadratic bound. Replacing $J_2(0)$ by $J_3(0)$ preserves the optimal value $a^2$ but makes it unattainable. These examples distinguish attainment by a distribution, uniform finite-step estimates, and attainment by a positive definite quadratic bound.

\subsection{Complete numerical records for attainable and unattainable cases}\label{app-toc-b-6}
\label{app:attainment-examples}

This section provides the full constructions and results used in the numerical illustrations in the main text, including support counts, sample ranks, and characteristic polynomials of the second-moment operators. Means, ranks, and characteristic polynomials of the finite-support constructions are computed in exact rational arithmetic.

\paragraph{Approaching an unattained infimum.}
For $A=\operatorname{diag}(a,J_3(0))$ with $a=0.6$ and rank budget $r=1$, the optimal factor is $q_1(A)=a^2=0.36$. Here $b(A)=1$ and $\nu_0(A)=3$, so Theorem~\ref{thm:attainment-appendix} rules out attainment.

For $\varepsilon>0$, set $c_\varepsilon=a^2+2\varepsilon^2$. Among the three nonzero rows of $A$, select the row containing $a$ with probability $a^2/c_\varepsilon$ and each of the two remaining nonzero rows in the zero chain with probability $\varepsilon^2/c_\varepsilon$. Retain only the selected row and divide it by its selection probability. The three support matrices and probabilities are
\[
\begin{array}{c|c}
\text{Matrix}&\text{Probability}\\\hline
\dfrac{c_\varepsilon}{a}e_1e_1^\top
&\dfrac{a^2}{c_\varepsilon}\\
\dfrac{c_\varepsilon}{\varepsilon^2}e_2e_3^\top
&\dfrac{\varepsilon^2}{c_\varepsilon}\\
\dfrac{c_\varepsilon}{\varepsilon^2}e_3e_4^\top
&\dfrac{\varepsilon^2}{c_\varepsilon}
\end{array}
\]
The probabilities sum to one, every sample has rank one, and the mean is exactly $A$. The second-moment operator maps every matrix to a diagonal matrix. Its representation on the diagonal subspace is
\[
\begin{pmatrix}
c_\varepsilon&0&0&0\\
0&0&c_\varepsilon/\varepsilon^2&0\\
0&0&0&c_\varepsilon/\varepsilon^2\\
0&0&0&0
\end{pmatrix}.
\]
This is upper triangular with diagonal $(c_\varepsilon,0,0,0)$, so
\[
\rho(\mathcal M_\varepsilon)=c_\varepsilon,
\;
\mathbb E\|R\|_F^2
=c_\varepsilon+\frac{2c_\varepsilon}{\varepsilon^2}.
\]
On the full space $\operatorname{Sym}^4$, the characteristic polynomial is $z^9(z-c_\varepsilon)$. Table~\ref{tab:app-b-nonattainment} lists the results at six scales.

\begin{table}[htbp]
\centering
\caption{Six sampling scales approaching the unattained optimal value $0.36$. Each distribution has three support matrices.}
\label{tab:app-b-nonattainment}
\small
\begin{tabular}{@{}rrrr@{}}
\toprule $\varepsilon$ & $\rho(\mathcal M_\varepsilon)$ & $\rho(\mathcal M_\varepsilon)-0.36$
& $\mathbb E\|R\|_F^2$\\
\midrule
0.40 & 0.6800 & 0.3200 & 9.1800\\
0.20 & 0.4400 & 0.0800 & 22.4400\\
0.10 & 0.3800 & 0.0200 & 76.3800\\
0.05 & 0.3650 & 0.0050 & 292.3650\\
0.02 & 0.3608 & 0.0008 & 1804.3608\\
0.01 & 0.3602 & 0.0002 & 7204.3602\\
\bottomrule
\end{tabular}
\end{table}

In this family, the gap to the optimal factor is $2\varepsilon^2$, while the one-step second moment $\mathbb E\|R\|_F^2$ grows on the order of $\varepsilon^{-2}$. The gap decreases by sampling the nonzero rows of the zero chain less often and applying larger unbiased compensating weights when they are selected. Every positive $\varepsilon$ yields an admissible distribution with finite second moment; the growth of second moments in the table describes this particular family.

\paragraph{Eight finite-support optimal constructions.}
We implement the construction in Section~\ref{app:attainment-constructions} for eight mean matrices of dimensions four to seven. The cases include multiple nonzero real eigenvalues, a nonreal conjugate pair, nontrivial Jordan chains at real and nonreal eigenvalues, multiple zero chains, equality of the two lower bounds, strict dominance of the spectral-radius term, and a nilpotent mean. Write
\[
C=\begin{pmatrix}1&-1\\1&1\end{pmatrix},
\;
B=\begin{pmatrix}1&3\\0&1\end{pmatrix},
\;
H=\begin{pmatrix}C&I_2\\0&C\end{pmatrix}.
\]
Table~\ref{tab:app-b-attainment-cases} gives the means up to similarity; the constructions use the interleaved coordinate order of Section~\ref{app:attainment-constructions}. The column $N$ counts entries in the finite sampling list, the sample-rank column lists every rank occurring in the distribution, and $\chi_{\mathcal M}$ is the characteristic polynomial on $\operatorname{Sym}^d$. In every row, the probabilities sum to one, $\mathbb ER=A$, and $\rho(\mathcal M)=q_r(A)$.

\begin{table}[htbp]
\centering
\caption{Complete data for eight finite-support optimal distributions. The sample-rank column lists all ranks occurring in each distribution.}
\label{tab:app-b-attainment-cases}
\small
\setlength{\tabcolsep}{5pt}
\begin{tabular}{@{}lrrrrcl@{}}
\toprule
Mean, up to similarity & $d$ & $r$ & $N$ & $q_r(A)$
& Sample ranks & $\chi_{\mathcal M}(z)$\\
\midrule $\operatorname{diag}(1,2,J_3(0))$ & 5 & 1 & 4 & 5 & $1$ & $z^{13}(z-5)^2$\\ $\operatorname{diag}(C,J_2(0))$ & 4 & 1 & 4 & 4 & $1$ & $z^9(z-4)$\\ $\operatorname{diag}(1,1,J_3(0))$ & 5 & 2 & 4 & 1 & $2$ & $z^{12}(z-1)^3$\\ $\operatorname{diag}(B,J_3(0))$ & 5 & 1 & 4 & 2 & $1$ & $z^{13}(z-2)^2$\\ $\operatorname{diag}(H,J_3(0))$ & 7 & 2 & 8 & 4 & $2$ & $z^{25}(z-4)^3$\\ $\operatorname{diag}(1,1,J_3(0),J_2(0))$ & 7 & 2 & 8 & 1 & $2$ & $z^{25}(z-1)^3$\\ $\operatorname{diag}(1,\tfrac12,J_4(0))$ & 6 & 2 & 6 & 1 & $1,2$ & $z^{18}(z-1)^2(z-\tfrac12)$\\ $J_4(0)$ & 4 & 1 & 4 & 0 & $0,1$ & $z^{10}$\\
\bottomrule
\end{tabular}
\end{table}

The first six cases use the construction that interleaves zero layers and nonzero spectral blocks. The third and sixth satisfy $\mathcal E(A)/r=\rho(A)^2$, illustrating that the zero-chain length criterion also applies when the two lower bounds coincide. The fourth and fifth retain Jordan links at nonzero eigenvalues. The seventh satisfies $\rho(A)^2>\mathcal E(A)/r$, so an optimal distribution exists even though $\nu_0(A)=4>b(A)+1=3$. In the final case, products vanish after finitely many steps and the growth factor is zero.

A neighboring unattainable example is $A=\operatorname{diag}(C,J_3(0))$ with $r=1$. Here $b(A)=1$ and $\nu_0(A)=3$, exceeding the permitted length $b(A)+1=2$. The polynomial $p(z)=z^2(z^2-2z+2)$ does not annihilate this mean: on the nonzero block, $C^2-2C+2I=0$, whereas on the zero block, $p(J_3(0))=2J_3(0)^2\ne0$. This calculation displays the structural difference caused by the longer zero chain; nonattainment follows from Theorem~\ref{thm:attainment-appendix}.

\section{One-step error, finite-time bounds, and temporal dependence}\label{app-toc-c}
\label{app:extensions}

The optimal value in the main text describes asymptotic mean-square growth under independent repetition of a fixed sampling law. This appendix considers three related questions in turn: how accurately a single random matrix can approximate its mean, what finite-time bounds are available under a condition-number budget, and how the optimal growth changes when sampling laws vary over time or depend on the history.

\subsection{One-step moments, approximation error, and optimal laws}\label{app-toc-c-1}
\label{app:one-step}

For a rectangular mean $B\in\mathbb R^{m\times n}$, the class $\mathcal C_r(B)$ is defined by the same requirements as in the main text: $\mathbb E R=B$, $\operatorname{rank} R\le r$ almost surely, and $\mathbb E\|R\|_F^2<\infty$, where $r$ is an integer with $1\le r\le\min(m,n)$. The largest directional second moment is $\sup_{\|x\|_2=1}\mathbb E\|Rx\|_2^2=\lambda_{\max}(\mathbb E[R^{\top}R])$. Replacing $R$ by $R-B$ gives the largest directional approximation error. The general quadratic budget criterion in Proposition~\ref{thm:budget} solves both problems.

\subsubsection{The largest directional second moment}
\label{app:raw-moment}

\begin{corollary}[Optimal one-step second moment]
\label{cor:raw}
For every $B\in\mathbb R^{m\times n}$,
\begin{equation}
\label{eq:app-c-raw}
\gamma_r(B):=\inf_{\mu\in\mathcal C_r(B)}\lambda_{\max}(\mathbb E[R^{\top}R])
=\max\left\{\|B\|_2^2,\frac{\|B\|_F^2}{r}\right\}.
\end{equation}
A finite-support law attains this value. If $B\ne0$ has compact singular value decomposition $B=\sum_{i=1}^{s}\sigma_i u_iv_i^{\top}$, one such law is
\begin{equation}
\label{eq:app-c-raw-sampler}
R=\sum_{i\in S}\frac{\sigma_i}{p_i}u_iv_i^{\top},
\; p_i=\frac{\sigma_i^2}{\gamma_r(B)},
\end{equation}
where $\Pr(i\in S)=p_i$ and $|S|\le r$ as in Lemma~\ref{lem:subset}. It satisfies $\mathbb E[R^{\top}R]=\gamma_r(B)\Pi_{\mathrm{row}(B)}$, where $\Pi_{\mathrm{row}(B)}$ is the orthogonal projection onto the row space of $B$.
\end{corollary}

\begin{proof}
For $c>0$, Proposition~\ref{thm:budget} with $P=I_m$ and $Q=cI_n$ says that $\mathbb E[R^{\top}R]\preceq cI_n$ is feasible exactly when $c\ge\|B\|_2^2$ and $rc\ge\|B\|_F^2$. This proves the lower bound. For $B\ne0$, the probabilities in \eqref{eq:app-c-raw-sampler} satisfy $0<p_i\le1$ and $\sum_i p_i\le r$. Lemma~\ref{lem:subset} supplies a finite-support subset law. Orthogonality of the left singular vectors gives
\[
\mathbb E[R^{\top}R]=\sum_{i=1}^{s}\frac{\sigma_i^2}{p_i}v_iv_i^{\top}
=\gamma_r(B)\Pi_{\mathrm{row}(B)}.
\]
The mean is $B$ and the rank is at most $r$, proving attainment. For $B=0$, take $R=0$.
\end{proof}

For $B\ne0$, its stable rank is $\operatorname{srank}(B):=\|B\|_F^2/\|B\|_2^2$. Thus $\gamma_r(B)=\|B\|_2^2$ exactly when $r\ge\operatorname{srank}(B)$: this is the rank budget needed to match the largest directional second moment of the deterministic matrix. Within singular-component samplers, the probabilities in \eqref{eq:app-c-raw-sampler} equalize all ratios $\sigma_i^2/p_i$. Proposition~\ref{thm:budget} establishes their optimality over the full class $\mathcal C_r(B)$.

\subsubsection{The largest directional approximation error}
\label{app:centered-error}

\begin{corollary}[Optimal one-step approximation error]
\label{cor:centered}
Let $s=\operatorname{rank} B$, with positive singular values $\sigma_1,\ldots,\sigma_s$, and define
\[
\tau_r(B):=\inf_{\mu\in\mathcal C_r(B)}
\lambda_{\max}\bigl(\mathbb E[(R-B)^{\top}(R-B)]\bigr).
\]
If $r\ge s$, then $\tau_r(B)=0$. If $r<s$, then $\tau_r(B)$ is the unique positive solution of
\begin{equation}
\label{eq:app-c-tau}
\sum_{i=1}^{s}\frac{\sigma_i^2}{\sigma_i^2+\tau_r(B)}=r.
\end{equation}
In the latter case, the sampler \eqref{eq:app-c-raw-sampler} with
\begin{equation}
\label{eq:app-c-centered-sampler}
p_i=\frac{\sigma_i^2}{\sigma_i^2+\tau_r(B)}
\end{equation}
attains the optimum. It selects exactly $r$ components almost surely and satisfies $\mathbb E[(R-B)^{\top}(R-B)]=\tau_r(B)\Pi_{\mathrm{row}(B)}$.
\end{corollary}

\begin{proof}
Unbiasedness gives \(\mathbb E[(R-B)^{\top}(R-B)]=\mathbb E[R^{\top}R]-B^{\top}B\). For $t>0$, an error bound of $t$ is therefore equivalent to $\mathbb E[R^{\top}R]\preceq B^{\top}B+tI_n$. Apply Proposition~\ref{thm:budget} with $P=I_m$ and $Q=B^{\top}B+tI_n$. Its first condition is automatic; its second is
\[
\operatorname{tr}\bigl((B^{\top}B+tI_n)^{-1}B^{\top}B\bigr)
=\sum_{i=1}^{s}\frac{\sigma_i^2}{\sigma_i^2+t}\le r.
\]
When $r<s$, this sum decreases continuously and strictly from $s$ to zero, so the smallest feasible positive bound is the solution of \eqref{eq:app-c-tau}. The probabilities in \eqref{eq:app-c-centered-sampler} sum to the integer $r$. Lemma~\ref{lem:subset} then gives $|S|=r$ almost surely, and
\[
\mathbb E[R^{\top}R]=\sum_{i=1}^{s}\bigl(\sigma_i^2+\tau_r(B)\bigr)v_iv_i^{\top}.
\]
Subtracting $B^{\top}B$ proves attainment. When $r\ge s$, the deterministic choice $R=B$ has zero error.
\end{proof}

Matching the deterministic largest second moment requires $r\ge\operatorname{srank}(B)$, whereas zero approximation error requires $r\ge\operatorname{rank} B$. For $B=\beta I_d$, the formulas become $\gamma_r(B)=\beta^2d/r$ and $\tau_r(B)=\beta^2(d/r-1)$. For unequal singular values, the centered optimum depends on the entire singular spectrum through \eqref{eq:app-c-tau}.

\subsubsection{Equality and all optimal laws}
\label{app:optimizer-structure}

The next identity identifies what equality in the rank bound requires. It also describes all second-moment optimizers for an invertible mean when the Frobenius bound determines the optimum.

\begin{lemma}[Equality in the rank--moment bound]
\label{lem:defect}
Let $Y\in\mathbb R^{m\times n}$ satisfy $\mathbb E Y=D$, $\operatorname{rank} Y\le r$ almost surely, and $C:=\mathbb E[Y^{\top}Y]\preceq I_n$. If $\Pi$ is the orthogonal projection onto $\operatorname{range}(Y)$, then
\begin{equation}
\label{eq:app-c-defect}
r-\|D\|_F^2=\mathbb E[r-\operatorname{rank} Y]+\mathbb E\|\Pi-YD^{\top}\|_F^2
+\operatorname{tr}\bigl(D(I_n-C)D^{\top}\bigr).
\end{equation}
Each term on the right is nonnegative. If $\|D\|_F^2=r$, then $\operatorname{rank} Y=r$ and $\Pi=YD^{\top}$ almost surely, $(I_n-C)D^{\top}=0$, and $\mathbb E\Pi=DD^{\top}$.
\end{lemma}

\begin{proof}
Since $\Pi Y=Y$,
\[
\mathbb E\|\Pi-YD^{\top}\|_F^2
=\mathbb E[\operatorname{rank} Y]-2\|D\|_F^2+\operatorname{tr}(DCD^{\top}).
\]
Rearrangement gives \eqref{eq:app-c-defect}. If its left side vanishes, every term on the right vanishes. The last term equals $\|(I_n-C)^{1/2}D^{\top}\|_F^2$, which gives $(I_n-C)D^{\top}=0$. Taking expectations in $\Pi=YD^{\top}$ gives $\mathbb E\Pi=DD^{\top}$.
\end{proof}

\begin{proposition}[All one-step optimizers for an invertible mean]
\label{prop:optimizers}
Suppose $B\in\mathbb R^{d\times d}$ is invertible and $\gamma:=\|B\|_F^2/r\ge\|B\|_2^2$. A law in $\mathcal C_r(B)$ attains \eqref{eq:app-c-raw} if and only if
\begin{equation}
\label{eq:app-c-all-optimizers}
R=\gamma\Pi B^{-\top},\;
\Pi^{\top}=\Pi=\Pi^2,\;
\operatorname{rank}\Pi=r\ \text{a.s.},\;
\mathbb E\Pi=\frac{BB^{\top}}{\gamma}.
\end{equation}
Every such law satisfies $\mathbb E[R^{\top}R]=\gamma I_d$.
\end{proposition}

\begin{proof}
For an optimal law, take $Y=R/\sqrt\gamma$ and $D=B/\sqrt\gamma$. Then $\mathbb E[Y^{\top}Y]\preceq I_d$ and $\|D\|_F^2=r$. Lemma~\ref{lem:defect} gives $\operatorname{rank}\Pi=r$ and $\Pi=RB^{\top}/\gamma$ almost surely, together with $\mathbb E\Pi=BB^{\top}/\gamma$. Invertibility of $B$ yields \eqref{eq:app-c-all-optimizers}.

Conversely, \eqref{eq:app-c-all-optimizers} implies $\mathbb E R=B$, $\operatorname{rank} R=r$ almost surely, and \(\mathbb E[R^{\top}R]=\gamma^2 B^{-1}(\mathbb E\Pi)B^{-\top}=\gamma I_d\). The samples are uniformly bounded because $\|\Pi\|_2=1$, so the law has finite second moment and attains the optimum.
\end{proof}

The prescribed expectation obeys $0\preceq BB^{\top}/\gamma\preceq I_d$ and $\operatorname{tr}(BB^{\top}/\gamma)=r$. Diagonalizing it and applying Lemma~\ref{lem:subset} constructs a random rank-$r$ projection with this mean. Thus \eqref{eq:app-c-all-optimizers} parameterizes the optimal laws by their random range projections. Singular-component sampling uses projections onto subsets of the left singular vectors; other projection laws give other optimizers.

\subsection{One-step criteria and repeated sampling}\label{app-toc-c-2}
\label{app:one-step-examples}

\subsubsection{Identical one-step moments, different dynamics}
\label{app:identical-moments}

Section~3 compares three laws with mean $A=\operatorname{diag}(0.9,0.4)$ and rank budget $r=1$. Here we give the full second-moment calculations, showing how repeated application retains directional information that the one-step criteria do not distinguish.
The spectral law selects
\[
 R=\begin{cases}
 \operatorname{diag}(1.3,0),&\text{with probability }9/13,\\
 \operatorname{diag}(0,1.3),&\text{with probability }4/13,
 \end{cases}
\]
whereas the dense law assigns equal probabilities to
\[
 \begin{pmatrix}0.9&0.6\\0.6&0.4\end{pmatrix},
 \;
 \begin{pmatrix}0.9&-0.6\\-0.6&0.4\end{pmatrix}.
\]
Both laws have mean $A$, and every realization has rank one.
Their full Euclidean moment matrices agree:
\[
 \mathbb E[R^\top R]=\operatorname{diag}(1.17,0.52),
 \;
 \mathbb E[(R-A)^\top(R-A)]=0.36I_2.
\]
Thus both have Frobenius mean square error $0.72$ and attain the
centered directional optimum $\tau_1(A)=0.36$ from Corollary~\ref{cor:centered}.

The Frobenius error is also globally minimal. For any matrix of rank
at most one, $|\operatorname{tr} R|\le\|R\|_F$. Therefore
\[
 \mathbb E\|R\|_F^2\ge(\mathbb E\operatorname{tr} R)^2=(\operatorname{tr} A)^2=1.69.
\]
Unbiasedness gives
$\mathbb E\|R-A\|_F^2=\mathbb E\|R\|_F^2-\|A\|_F^2$.
Since $\|A\|_F^2=0.97$, the Frobenius error is at least $0.72$.
Both laws attain this lower bound, giving a concrete example of nonuniqueness in unbiased Frobenius-optimal sampling as studied by \citet{benzing2019} and \citet{barnes2025}.

Their moment operators contain different information. In coordinates
$(X_{11},X_{12},X_{22})$ on $\operatorname{Sym}^2$, they are represented by
\[
 M_{\rm sp}=\begin{pmatrix}
 1.17&0&0\\0&0&0\\0&0&0.52
 \end{pmatrix},
 \;
 M_{\rm de}=\begin{pmatrix}
 0.81&0&0.36\\0&0.72&0\\0.36&0&0.16
 \end{pmatrix}.
\]
Consequently, $\rho(M_{\rm sp})=1.17$, whereas
$\rho(M_{\rm de})=0.97=q_1(A)$.
The dense law also satisfies
$\mathbb E[R^\top PR]=0.97P$ with $P=A\succ0$.
The two laws give the same expected squared length after one application, but their second moments propagate differently at subsequent steps.

For comparison, the raw-optimal sampler in Corollary~\ref{cor:raw} selects
$\operatorname{diag}(97/90,0)$ and $\operatorname{diag}(0,97/40)$ with probabilities $81/97$
and $16/97$, respectively. It satisfies $\mathbb E[R^\top R]=0.97I_2$
and has asymptotic factor $0.97$, with larger centered and Frobenius
errors. Table~\ref{tab:app-c-laws} records all four quantities for the three laws.

\begin{table}[tbp]
\centering
\caption{Three unbiased rank-one laws with mean $\operatorname{diag}(0.9,0.4)$.
The first two share the entire right second-moment matrix.
The centered and raw columns give the largest eigenvalues of the
corresponding moment matrices.}
\label{tab:app-c-laws}\label{tab:app-c-three-laws}
\small
\setlength{\tabcolsep}{5pt}
\begin{tabular}{@{}lrrrr@{}}
\toprule
Law & Frobenius MSE & Centered moment & Raw moment & $g(\mu)$\\
\midrule
Spectral & $0.72$ & $0.36$ & $1.17$ & $1.17$\\
Dense & $0.72$ & $0.36$ & $1.17$ & $0.97$\\
Raw optimal & $0.97$ & $0.81$ & $0.97$ & $0.97$\\
\bottomrule
\end{tabular}
\end{table}

\begin{figure}[tbp]
\centering
\includegraphics[width=.8\linewidth]{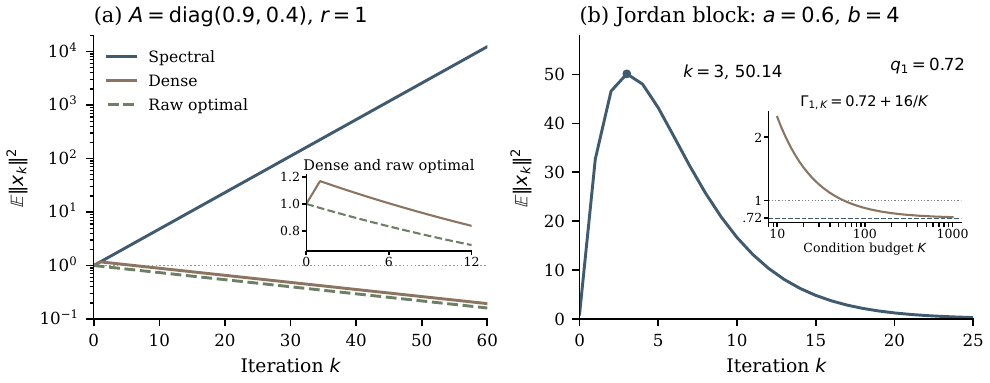}
\caption{Second moments computed by deterministic recursions.
Left: the three laws in Table~\ref{tab:app-c-laws}, starting from $e_1$, with an
inset showing the initial behavior of the two stable laws.
Right: the Jordan law \eqref{eq:app-c-Jordan-law} with $a=0.6$ and $b=4$,
starting from $e_2$; the inset shows $\Gamma_{1,K}=0.72+16/K$.
All expectations are evaluated over the complete finite support.}
\label{fig:app-c-examples}
\end{figure}

The spectral law also separates almost sure behavior from mean square
behavior. Starting from $x_0=e_1$, the product is nonzero after $k$
steps only if the first coordinate was selected in every draw. That
event has probability $(9/13)^k$, and the squared amplitude on it is
$1.69^k$. Hence $\mathbb E\|x_k\|_2^2=1.17^k$, even though the state reaches
zero in finite time almost surely. The growing second moment comes
from increasingly rare surviving trajectories.
Figure~\ref{fig:app-c-examples} shows the exact second-moment trajectories for all
three laws. Its right panel displays the transient amplification of
the explicit Jordan law \eqref{eq:app-c-Jordan-law}, analyzed below.

\subsubsection{Directional criteria across singular spectra}
\label{app:singular-spectrum-experiment}
The sampling probabilities show how each objective distributes the
rank budget among singular directions. We compare three families of
$15\times12$ matrices with singular values $\sigma_i=1$,
$\sigma_i=0.8^{i-1}$, and
$(\sigma_1,\ldots,\sigma_{12})=(1,0.2,\ldots,0.2)$.
The left singular vectors are the first twelve columns of the
orthogonal cosine matrix of order fifteen; the right singular
vectors form the orthogonal cosine matrix of order twelve.
Thus every target is dense and rectangular. The stable ranks of
the flat, geometric, and spiked spectra are $12$, $2.7647$, and
$1.44$, respectively.

For each $r=1,\ldots,12$, we construct laws minimizing the raw
directional moment, the centered directional error, and the
Frobenius mean square error. All three use singular-component
sampling and have rank exactly $r$ in every realization.
For $r<12$, the raw construction uses \(p_i=\min\{1,\sigma_i^2/t\},\; \sum_i p_i=r\), where $t>0$ is chosen to satisfy the sum constraint. These
probabilities also attain the optimum in Corollary~\ref{cor:raw}, because \(\max_i\frac{\sigma_i^2}{p_i} =\max\{\sigma_1^2,t\}=\gamma_r(B)\). This choice uses the full rank budget even above the stable-rank
threshold. The Frobenius construction uses
$p_i=\min\{1,\sigma_i/t\}$ with $\sum_i p_i=r$
\citep{benzing2019,barnes2025}, and the centered construction uses
\eqref{eq:app-c-centered-sampler}. The subset construction in
Lemma~\ref{lem:subset} enforces exact cardinality in each case.
At $r=12$, all three laws equal the target deterministically.

\begin{figure}[tbp]
\centering
\includegraphics[width=.9\linewidth]{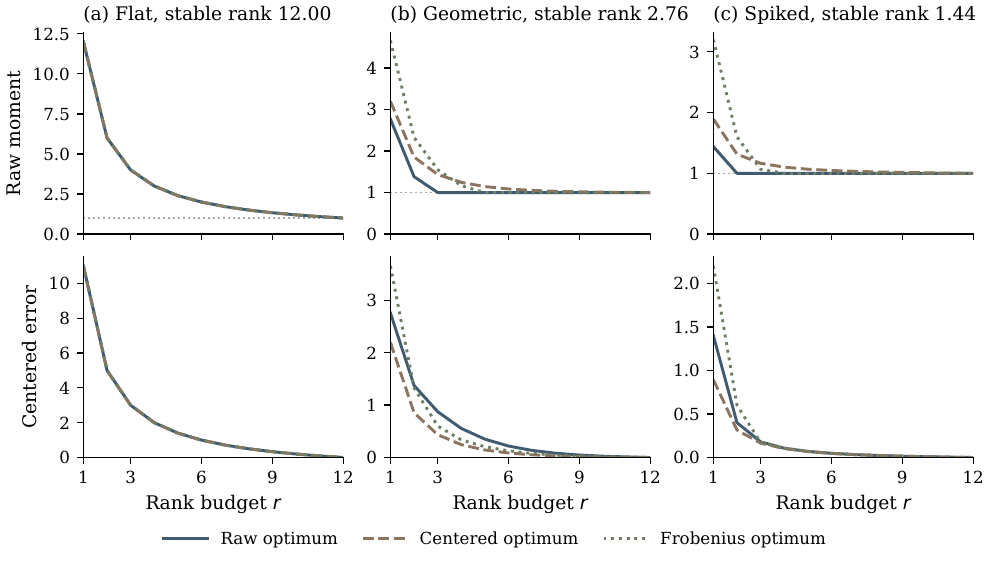}
\caption{Directional moments of dense $15\times12$ targets with
three singular spectra. Each curve is evaluated over the complete
support of its law, with every realization of rank exactly $r$.
The top row gives $\lambda_{\max}(\mathbb E[R^\top R])$ and the bottom
row gives $\lambda_{\max}(\mathbb E[(R-B)^\top(R-B)])$.}
\label{fig:app-c-spectra}
\end{figure}

Figure~\ref{fig:app-c-spectra} shows all rank budgets for the three spectra.
The constructions coincide for the flat spectrum and separate
when the singular values are unequal. For the geometric spectrum
at rank two, their raw, centered, and Frobenius values are listed
in Table~\ref{tab:app-c-geometric-spectrum}. Each construction minimizes its
designated objective, but these three optimizers allocate the
rank budget differently.

\begin{table}[tbp]
\centering
\caption{One-step criteria for the geometric-spectrum $15\times12$
target at rank two. Each row minimizes its named objective.}
\label{tab:app-c-geometric-spectrum}
\small
\begin{tabular}{@{}lrrr@{}}
\toprule
Sampling objective & Raw moment & Centered moment & Frobenius MSE\\
\midrule
Raw moment & $1.3823$ & $1.3750$ & $13.8233$\\
Centered moment & $1.8544$ & $0.8544$ & $10.2524$\\
Frobenius MSE & $2.3282$ & $1.3282$ & $8.0764$\\
\bottomrule
\end{tabular}
\end{table}

The raw optimum reaches $1$ at rank three for the geometric
spectrum and rank two for the spiked spectrum, as their stable
ranks predict. The centered error remains positive until rank
twelve. Matching the deterministic raw moment therefore requires
less rank than eliminating the approximation error.

\subsection{Conditioning and finite-time bounds}\label{app-toc-c-3}
\label{app:conditioning}

The estimate $\mathbb E\|x_k\|_2^2\le\kappa_2(P)c^k\|x_0\|_2^2$ controls both exponential growth and its prefactor. A factor close to $q_r(A)$ can require a poorly conditioned quadratic form, which may weaken the estimate at finite times. To quantify the tradeoff, fix $K\ge1$ and define
\[
\Gamma_{r,K}(A):=
\inf_{\substack{\mu\in\mathcal C_r(A),\ P\succ0\\\kappa_2(P)\le K}}
\lambda_{\max}\bigl(P^{-1/2}\mathcal M_\mu^*(P)P^{-1/2}\bigr).
\]
Here $\mathcal M_\mu^*(P)=\mathbb E[R^\top PR]$.

\subsubsection{Optimization under a condition-number budget}
\label{app:condition-budget}

\begin{proposition}[A condition-number budget]
\label{prop:condition}
For every $K\ge1$,
\begin{equation}
\label{eq:app-c-Gamma}
\Gamma_{r,K}(A)=
\min_{I_d\preceq P\preceq KI_d}
\max\left\{\|P^{1/2}AP^{-1/2}\|_2^2,
\frac{\|P^{1/2}AP^{-1/2}\|_F^2}{r}\right\}.
\end{equation}
The minimum is attained by a positive definite matrix and a finite-support law. For this law,
\begin{equation}
\label{eq:app-c-condition-finite-time}
\mathbb E\|x_k\|_2^2\le K\Gamma_{r,K}(A)^k\|x_0\|_2^2,
\; k\ge1.
\end{equation}
Moreover, $\Gamma_{r,1}(A)=\gamma_r(A)$ and $\Gamma_{r,K}(A)\downarrow q_r(A)$ as $K\to\infty$.
\end{proposition}

\begin{proof}
For fixed $P$, apply Corollary~\ref{cor:raw} to $P^{1/2}RP^{-1/2}$, whose mean is $P^{1/2}AP^{-1/2}$. Multiplying $P$ by a positive scalar changes neither the resulting objective nor its condition number, so $\kappa_2(P)\le K$ can be normalized to $I_d\preceq P\preceq KI_d$. The normalized set is compact and the objective is continuous. A minimizer therefore exists, and Corollary~\ref{cor:raw} supplies a finite-support law that realizes it. Lemma~\ref{lem:lyapunov} gives \eqref{eq:app-c-condition-finite-time}.

At $K=1$, the normalized feasible set contains only $I_d$. As $K$ increases, these sets exhaust the positive definite matrices up to scaling. The metric variational formula in Section~4 gives the asserted monotone limit.
\end{proof}

\subsubsection{Sharp conditioning exponents}
\label{app:conditioning-exponents}

The rate of approach to $q_r(A)$ depends on Jordan chains. The Frobenius constraint detects defective eigenvalues throughout the spectrum; the operator norm constraint detects defective eigenvalues of maximal modulus. Their respective costs determine the following exponent.

\begin{theorem}[Sharp conditioning exponents]
\label{thm:conditioning-exponents}
Write $u:=\mathcal E(A)/r$, $v:=\rho(A)^2$, and $q:=\max\{u,v\}$. Let $s$ be the largest Jordan block size of $A$, minus one. If $\rho(A)>0$, let $p$ be the largest Jordan block size at an eigenvalue of modulus $\rho(A)$, minus one. All block sizes are taken over $\mathbb C$. For $A\ne0$, define
\[
\alpha:=
\begin{cases}
s,&q=0,\\
s,&u>v,\\
\max\{s,2p\},&u=v>0,\\
2p,&v>u.
\end{cases}
\]
If $\alpha>0$, there are $c,C,K_0>0$, depending only on $A,r$, such that
\begin{equation}
\label{eq:app-c-conditioning-power-law}
cK^{-1/\alpha}\le\Gamma_{r,K}(A)-q\le CK^{-1/\alpha},
\; K\ge K_0.
\end{equation}
If $\alpha=0$, then $\Gamma_{r,K}(A)=q$ for all sufficiently large $K$. If $A=0$, equality holds for every $K\ge1$. The upper bounds are realized by finite-support laws.
\end{theorem}

\begin{proof}
Polar decomposition identifies the singular values of $SAS^{-1}$ with those of $P^{1/2}AP^{-1/2}$ for $P=S^{\top}S$. Also $\kappa_2(P)=\kappa_2(S)^2$. We may therefore optimize over real similarities $B=SAS^{-1}$ with $\kappa_2(S)\le\sqrt K$.

\emph{Frobenius lower bound.} Suppose $s\ge1$ and $\|B\|_2\le M$ for a fixed $M$. Let $\chi(z):=\prod_{\lambda\in\sigma(A)}(z-\lambda)$, with each distinct eigenvalue included once. Since its roots are simple, $\chi(A)$ is nilpotent of index $s+1$, and $\chi(A)^s\ne0$. A complex Schur form $U^*BU=D+T$, with $D$ diagonal and $T$ strictly upper triangular, satisfies \(\chi(D)=0,\; \|T\|_F^2=\|B\|_F^2-\mathcal E(A)\). The identity $X^j-Y^j=\sum_{\ell=0}^{j-1}X^{j-1-\ell}(X-Y)Y^\ell$ applied termwise to $\chi$ gives $\|\chi(B)\|_2\le L\|T\|_F$, where $L$ depends only on $\chi,M$. Consequently,
\[
L^s\|T\|_F^s\ge\|\chi(B)^s\|_2
=\|S\chi(A)^sS^{-1}\|_2
\ge\frac{\|\chi(A)^s\|_2}{\sqrt K}.
\]
For some $c_1>0$,
\begin{equation}
\label{eq:app-c-departure-lower-conditioning}
\|B\|_F^2-\mathcal E(A)\ge c_1K^{-1/s}.
\end{equation}

\emph{Operator norm lower bound.} Suppose $\rho:=\rho(A)>0$ and $p\ge1$. A peripheral Jordan chain gives $\|A^k\|_2\ge c_2k^p\rho^k$ for all sufficiently large $k$: its corner entry is $\binom{k}{p}\lambda^{k-p}$, and the fixed change of basis changes only $c_2$. Hence \(c_2k^p\rho^k\le\sqrt K\,\|B\|_2^k\). Set $\eta:=\|B\|_2/\rho-1$. A nontrivial peripheral Jordan block implies $\eta>0$, since $\|B\|_2\le\rho$ would bound $\rho^{-k}\|A^k\|_2$ uniformly. For small $\eta$, take $k=\lceil1/\eta\rceil$, so $(1+\eta)^k\le e^2$. The preceding inequality yields $\eta\ge c_3K^{-1/(2p)}$. For $\eta$ bounded away from zero, the same estimate follows after decreasing $c_3$. Thus
\begin{equation}
\label{eq:app-c-operator-lower-conditioning}
\|B\|_2^2-\rho^2\ge c_4K^{-1/(2p)}.
\end{equation}

At a minimizing metric in Proposition~\ref{prop:condition}, $\|B\|_2\le\sqrt{q+1}$ for all sufficiently large $K$, because $\Gamma_{r,K}(A)\downarrow q$. Thus \eqref{eq:app-c-departure-lower-conditioning} applies with a fixed $M$. It gives the desired lower bound when $u>v$ or $q=0$; \eqref{eq:app-c-operator-lower-conditioning} gives it when $v>u$. When $u=v>0$, both bounds apply, and the slower decay has exponent $1/\max\{s,2p\}$. An estimate with zero index is omitted.

\emph{Matching construction.} Fix a real Jordan representation $A=VJV^{-1}$. On a chain of length $\ell\ge2$, multiply successive coordinates by $t^{\ell-1},t^{\ell-2},\ldots,t,1$, repeating each scale twice for a conjugate-pair chain. Each Jordan link is then multiplied by $t$. Given $H\ge1$, use $t=H^{-1/(\ell-1)}$ on each chain. All diagonal entries of the resulting scaling $D_H$ lie in $[H^{-1},1]$, so $\kappa_2(D_H)\le H$. Take $S=D_HV^{-1}$ with $H=\sqrt K/\kappa_2(V)$. For sufficiently large $K$, this is feasible and, when $s\ge1$, \(\|SAS^{-1}\|_F^2-\mathcal E(A)=O(K^{-1/s})\). There is no linear cross term: the normal diagonal blocks and Jordan links occupy disjoint entries. If $s=0$, this excess is zero.

For $\rho>0$ and $p\ge1$, each peripheral block has norm at most $\rho+H^{-1/(\ell-1)}$, whereas each strictly interior block eventually has norm below $\rho$. Therefore \(\|SAS^{-1}\|_2^2-\rho^2=O(K^{-1/(2p)})\). If $p=0$, the left side is zero for sufficiently large $K$. These estimates prove the upper bound for $u\ge v>0$: when $u>v$, the operator norm constraint eventually has strict slack; when $u=v$, use both estimates. If $q=0$, all diagonal blocks vanish and both squared norms are $O(K^{-1/s})$.

When $v>u$, first scale the strictly interior chains by fixed amounts so that their operator norms are below $\rho$ and their total Frobenius excess is less than $(r\rho^2-\mathcal E(A))/2$. Allocate the remaining growing condition number budget to the peripheral chains. The fixed scaling changes only constants. The Frobenius constraint then has strict slack for sufficiently large $K$, and the operator norm estimate gives the exponent $1/(2p)$. If the relevant index is zero, a fixed finite scaling already satisfies both constraints at $q$.

Finally, apply Corollary~\ref{cor:raw} to $B=SAS^{-1}$. It supplies a finite-support law $\widetilde R$ with $\mathbb E[\widetilde R^{\top}\widetilde R]\preceq\gamma_r(B)I_d$. Setting $R=S^{-1}\widetilde R S$ and $P=S^{\top}S$ gives
\[
\mathbb E[R^{\top}PR]=S^{\top}\mathbb E[\widetilde R^{\top}\widetilde R]S\preceq\gamma_r(B)P,
\; \kappa_2(P)\le K.
\]
The mean, rank bound, and finite support are preserved.
\end{proof}

The distinction between the exponents has a direct source. A Jordan link of size $t$ adds a cost of order $t^2$ to the squared Frobenius norm. A link at a nonzero peripheral eigenvalue adds a cost of order $t$ to the squared operator norm. At $\mathcal E(A)/r=\rho(A)^2$, the slower decay of these two contributions determines the condition number requirement.

\begin{corollary}[Cost of a near-optimal quadratic bound]
\label{cor:conditioning-cost}
Under Theorem~\ref{thm:conditioning-exponents}, define
\[
K_{\min}(\varepsilon):=\inf\bigl\{\kappa_2(P):\ P\succ0,\ \mu\in\mathcal C_r(A),~
\mathcal M_\mu^*(P)\preceq(q+\varepsilon)P\bigr\}.
\]
If $\alpha>0$, then $K_{\min}(\varepsilon)=\Theta_{A,r}(\varepsilon^{-\alpha})$ as $\varepsilon\downarrow0$. If $\alpha=0$ or $A=0$, this quantity remains bounded.
\end{corollary}

\begin{proof}
Proposition~\ref{prop:condition} says that a bound with condition number at most $K$ exists exactly when $\Gamma_{r,K}(A)\le q+\varepsilon$. Inverting the two inequalities in \eqref{eq:app-c-conditioning-power-law} proves the claim.
\end{proof}

The quantity $K_{\min}$ measures the conditioning of quadratic certificates; the one-step second moment of a law and a uniform finite-step bound are separate quantities. The following examples make its relation to optimal laws explicit.

\subsubsection{Exact formulas for a single Jordan chain}
\label{app:chain-conditioning}

\begin{corollary}[Exact tradeoff for a Jordan chain]
\label{cor:chain-conditioning}
Let $A=aI_m+bN_m$, where $m\ge2$, $b\ne0$, and $N_m$ has ones on the first superdiagonal and zeros elsewhere. For every $K\ge1$,
\begin{align}
\Gamma_{1,K}(A)&=ma^2+(m-1)b^2K^{-1/(m-1)},
\label{eq:app-c-chain-rank-one}\\
\Gamma_{r,K}(bN_m)&=b^2\max\{1,(m-1)/r\}K^{-1/(m-1)},
\; 1\le r\le m.
\label{eq:app-c-chain-nilpotent}
\end{align}
If $a\ne0$ and $r<m$, then, for all sufficiently large $K$,
\begin{equation}
\label{eq:app-c-chain-intermediate-rank}
\Gamma_{r,K}(A)=\frac{ma^2}{r}+\frac{(m-1)b^2}{r}K^{-1/(m-1)}.
\end{equation}
If $a\ne0$ and $r=m$, then $\Gamma_{m,K}(A)-a^2=\Theta_{a,b,m}(K^{-1/[2(m-1)]})$.
\end{corollary}

\begin{proof}
Write $T=S(bN_m)S^{-1}$ with $\kappa_2(S)\le\sqrt K$. Since $\operatorname{tr} T=0$, we have $\|aI_m+T\|_F^2=ma^2+\|T\|_F^2$. In a unitary Schur form of $T$, the only potentially nonzero entry of its $(m-1)$st power is the upper-right entry, equal to the product of the first-superdiagonal entries. The arithmetic--geometric mean inequality gives
\[
\|T\|_F^2\ge(m-1)\|T^{m-1}\|_2^{2/(m-1)}
\ge(m-1)b^2K^{-1/(m-1)}.
\]
Also $\|T\|_2^{m-1}\ge\|T^{m-1}\|_2$ gives $\|T\|_2^2\ge b^2K^{-1/(m-1)}$. Both bounds are attained by $S=\operatorname{diag}(t^{m-1},t^{m-2},\ldots,t,1)$ with $t=K^{-1/[2(m-1)]}$, for which $T=btN_m$.

For $r=1$, the Frobenius term in \eqref{eq:app-c-Gamma} dominates, proving \eqref{eq:app-c-chain-rank-one}. For $a=0$, the same scaling attains both lower bounds and proves \eqref{eq:app-c-chain-nilpotent}. If $a\ne0$ and $r<m$, then $\|aI_m+btN_m\|_2^2\le(|a|+|b|t)^2<ma^2/r$ for sufficiently small $t$, proving \eqref{eq:app-c-chain-intermediate-rank}. The final claim follows from Theorem~\ref{thm:conditioning-exponents}, since $u=v=a^2$ and $s=p=m-1$.
\end{proof}

\subsubsection{A two-dimensional Jordan block and transient growth}
\label{app:jordan-transient}

\begin{proposition}[A two-dimensional Jordan block]\label{prop:jordan}
Let $A_b=\left(\begin{smallmatrix}a&b\\0&a\end{smallmatrix}\right)$,
where $a,b\in\mathbb R$. For rank budget $r=1$ and every $K\ge1$,
\begin{equation}\label{eq:app-c-jordan-Gamma}
 q_1(A_b)=2a^2,\;
 \Gamma_{1,K}(A_b)=2a^2+\frac{b^2}{K}.
\end{equation}
\end{proposition}

\begin{proof}
The optimal value follows from the formula in Section~4. For $r=1$, the
Frobenius term in \eqref{eq:app-c-Gamma} determines the objective. Write
$P=\left(\begin{smallmatrix}p&s\\s&t\end{smallmatrix}\right)\succ0$.
A direct expansion gives \(\operatorname{tr}(P^{-1}A_b^\top PA_b) =2a^2+b^2\frac{p^2}{pt-s^2}\). If the eigenvalues of $P$ are $m\le M$, then $p\ge m$ and
$pt-s^2=mM$. Consequently, $\kappa_2(P)\le K$ implies \(\frac{p^2}{pt-s^2}\ge\frac{m}{M}\ge\frac1K\). Equality is attained by $P=\operatorname{diag}(1,K)$, proving the formula.
\end{proof}

Thus when $ab\ne0$, no finite condition number gives an optimal quadratic bound. An optimal law nevertheless exists: with equal probabilities, take
\begin{equation}
\label{eq:app-c-Jordan-law}
R_1=\begin{pmatrix}2a&2b\\0&0\end{pmatrix},
\; R_2=\begin{pmatrix}0&0\\0&2a\end{pmatrix}.
\end{equation}
Both samples have rank at most one and their mean is $A_b$. For $X=\left(\begin{smallmatrix}u&v\\v&w\end{smallmatrix}\right)$,
\[
\mathcal M(X)=\begin{pmatrix}2a^2u+4abv+2b^2w&0\\0&2a^2w\end{pmatrix}.
\]
Its eigenvalues on $\operatorname{Sym}^2$ are $2a^2,2a^2,0$, so this law attains $q_1(A_b)$. For $a\ne0$, $x_0=e_2$, and $q=2a^2$, the second-moment recursion gives
\begin{equation}
\label{eq:app-c-Jordan-transient}
\mathbb E\|x_k\|_2^2=q^k+2b^2kq^{k-1},\; k\ge1.
\end{equation}
Indeed, after the first step the off-diagonal second moment is zero, the second diagonal entry is $q^k$, and the first obeys $u_{k+1}=qu_k+2b^2q^k$ with $u_1=2b^2$. For $b\ne0$, the ratio in \eqref{eq:app-c-Jordan-transient} to $q^k$ grows linearly, so this law has no bound $Cq^k$ with a time-independent constant.

For example, $a=0.6$ and $b=4$ give $q=0.72$, while $\mathbb E\|x_3\|_2^2=50.139648$. Equation~\eqref{eq:app-c-jordan-Gamma} also shows that a quadratic bound with $\kappa_2(P)\le K$ and factor $c<1$ exists exactly when $K>16/0.28$. The right panel of Figure~\ref{fig:app-c-examples} displays this transient behavior and the corresponding conditioning tradeoff. The asymptotic factor, the finite-time growth of a particular law, and the best condition number over all laws are therefore separately computable quantities.

The results distinguish three properties at the optimal factor. For a fixed law and $q=q_r(A)>0$, an optimal quadratic certificate implies a uniform bound $Cq^k$, which in turn implies attainment of $q$. Table~\ref{tab:app-c-endpoint} gives examples showing that neither implication can be reversed in general. In the middle row, Corollary~\ref{cor:rigid-rank-one} gives a uniform geometric bound for every optimal law, while Proposition~\ref{prop:quadratic-attainment} excludes an optimal positive definite quadratic bound.

\begin{table}[t]
\centering
\caption{Attainment and bounds at the optimal factor for rank one.
Here $a\ne0$, and $b\ne0$ in the last row.
The first two rows concern existence among admissible laws;
the last row concerns the specified law in
\eqref{eq:app-c-Jordan-law}.
A uniform bound means
$\mathbb E\|x_k\|_2^2\le Cq_1(A)^k\|x_0\|_2^2$
for all $x_0$ and $k\ge0$.}
\label{tab:app-c-endpoint}
\small
\setlength{\tabcolsep}{5pt}
\begin{tabular}{@{}lccc@{}}
\toprule
Mean or specified law
& Attainment
& Uniform bound
& Quadratic certificate\\
\midrule
$\operatorname{diag}(a,J_3(0))$
& No & No & No\\
$\operatorname{diag}(a,J_2(0))$
& Yes & Every optimal law & No\\
Law \eqref{eq:app-c-Jordan-law} for $A_b$
& Yes & No & No\\
\bottomrule
\end{tabular}
\end{table}

\subsubsection{The exact transition near a Jordan block}
\label{app:near-jordan}

A nonzero eigenvalue separation makes the optimal quadratic bound attainable with a finite condition number. The next formula quantifies this transition.

\begin{proposition}[Transition near a Jordan block]
\label{prop:near-jordan}
Let
\[
A=\begin{pmatrix}a&b\\0&a+\delta\end{pmatrix},\;
d_0:=\sqrt{b^2+\delta^2},\; K\ge1,
\]
where $a,b,\delta\in\mathbb R$. With $[z]_+:=\max\{z,0\}$, define
\[
w_K:=\frac12\left[\frac{d_0+|b|}{\sqrt K}-(d_0-|b|)\sqrt K\right]_+,
\; F_K:=a^2+(a+\delta)^2+w_K^2.
\]
Then
\begin{align}
\Gamma_{1,K}(A)&=F_K,
\label{eq:app-c-near-jordan-r1}\\
\Gamma_{2,K}(A)&=\frac{F_K+\sqrt{F_K^2-4a^2(a+\delta)^2}}2.
\label{eq:app-c-near-jordan-r2}
\end{align}
If $\delta\ne0$, each factor reaches its optimum $q_r(A)$ exactly when
\begin{equation}
\label{eq:app-c-near-jordan-threshold}
K\ge K_\star:=\left(\frac{\sqrt{b^2+\delta^2}+|b|}{|\delta|}\right)^2.
\end{equation}
For fixed $b\ne0$, $K_\star\sim4b^2/\delta^2$ as $\delta\to0$.
\end{proposition}

\begin{proof}
Set $c=a+\delta/2$ and $B=A-cI_2$. Its symmetric and skew-symmetric parts are
\[
H=\begin{pmatrix}-\delta/2&b/2\\b/2&\delta/2\end{pmatrix},
\; Z=\begin{pmatrix}0&b/2\\-b/2&0\end{pmatrix}.
\]
The eigenvalues of $H$ are $\pm h$, where $h=d_0/2$, and the absolute skew coefficient is $z=|b|/2$. In a singular value decomposition of a real similarity $S$, its left orthogonal factor does not change the norms of $SAS^{-1}$, and a scalar factor cancels. We may therefore write
\[
S=D_tV^{\top},\; D_t=\operatorname{diag}(e^{t/2},e^{-t/2}),
\; 0\le t\le\tfrac12\log K,
\]
with $V$ orthogonal. Orthogonal changes preserve the absolute skew coefficient; the off-diagonal entry $y$ of $V^{\top}HV$ ranges over $[-h,h]$. A possible sign of the skew coefficient does not affect its minimum absolute value. After conjugation by $D_t$, the skew coefficient becomes $z'=z\cosh t+y\sinh t$.

For a real traceless matrix $C$ similar to $B$, \(\|C\|_F^2=\operatorname{tr}(C^2)+4(z')^2=\delta^2/2+4(z')^2\). Restoring $cI_2$ and minimizing over $y$ gives
\[
\min_V\|D_tV^{\top}AVD_t^{-1}\|_F^2
=a^2+(a+\delta)^2+4[z\cosh t-h\sinh t]_+^2.
\]
Since $h\ge z\ge0$, the positive part decreases with $t$. Taking $t=\frac12\log K$ gives $F_K$ as the minimum squared Frobenius norm and proves \eqref{eq:app-c-near-jordan-r1}.

For $r=2$, the operator norm term in \eqref{eq:app-c-Gamma} dominates. The determinant of a similar matrix is $a(a+\delta)$, so if its squared Frobenius norm is $F$, its squared operator norm is
\[
\frac{F+\sqrt{F^2-4a^2(a+\delta)^2}}2.
\]
This is strictly increasing in $F$ on its domain. Thus the same similarity minimizes it, proving \eqref{eq:app-c-near-jordan-r2}. If $\delta\ne0$, reaching either optimum is equivalent to $w_K=0$. Solving this inequality gives \eqref{eq:app-c-near-jordan-threshold}.
\end{proof}

For $\delta=0$, equation~\eqref{eq:app-c-near-jordan-r1} reduces to \eqref{eq:app-c-jordan-Gamma}. For $\delta\ne0$, it also gives
\[
\Gamma_{1,K}(A)-q_1(A)
=\frac{(d_0+|b|)^2}{4K}\left[1-\frac K{K_\star}\right]_+^2.
\]
A diagonalizable matrix close to a defective one therefore retains the Jordan conditioning cost below a threshold of order $\delta^{-2}$.

\begin{figure}[tbp]
\centering
\includegraphics[width=.8\linewidth]{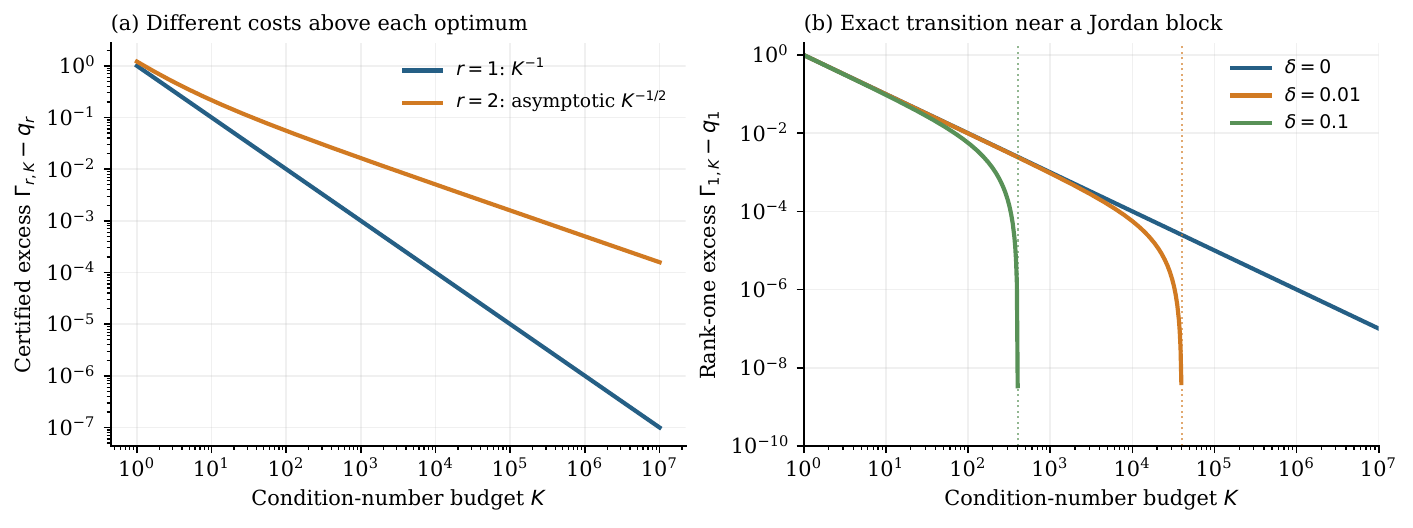}
\caption{Exact conditioning tradeoffs for
$A=\left(\begin{smallmatrix}0.5&1\\0&0.5+\delta\end{smallmatrix}\right)$.
Left: $\delta=0$, comparing the excess above each rank-dependent
optimum. Right: rank one and three eigenvalue separations; dotted
lines mark the finite cutoffs $K_\star$ beyond which the excess
is exactly zero. The zero portions are omitted on the logarithmic
scale.}
\label{fig:app-c-conditioning-transition}
\end{figure}

Figure~\ref{fig:app-c-conditioning-transition} plots the exact expressions in
Proposition~\ref{prop:near-jordan}. At $\delta=0$, the left panel compares
the excess above the two rank-dependent optima. Rank one and rank
two have different limiting values and different rates of approach,
as predicted by the two active moment constraints.
The right panel keeps rank one fixed and varies the eigenvalue
separation. For each nonzero separation, the excess reaches zero
at the finite threshold $K_\star$. As the eigenvalues approach
one another, this threshold increases, and the Jordan conditioning
cost remains visible over a longer range of condition numbers.

The accompanying computations optimize directly over positive
definite two-dimensional similarities for 16 parameter settings
and both rank budgets. Their values agree with the closed forms
to a maximum relative error of $1.3\times10^{-12}$.

\subsection{Independent schedules and sampling with memory}\label{app-toc-c-4}
\label{app:temporal}

\subsubsection{The optimum under independent time-varying laws}
\label{app:independent-schedules}

Allowing the sampling distribution to change at each step leaves the optimal growth factor unchanged, provided the draws remain independent. A finite-product lower bound gives this conclusion for every predetermined schedule.

\begin{theorem}[Independent time-varying sampling]
\label{thm:timevarying}
Let $(\mu_k)_{k\ge0}$ be a deterministic sequence of laws in $\mathcal C_r(A)$, and let $R_k\sim\mu_k$ be independent. Write $\mathcal M_k(X):=\mathbb E[R_kXR_k^{\top}]$ and $q:=q_r(A)$. For every integer $n\ge1$,
\begin{align}
\rho(\mathcal M_{n-1}\cdots\mathcal M_0)&\ge q^n,
\label{eq:app-c-operator-product-lower}\\
\sup_{\|x\|_2=1}\mathbb E\|R_{n-1}\cdots R_0x\|_2^2&\ge q^n.
\label{eq:app-c-finite-product-lower}
\end{align}
Consequently,
\begin{equation}
\label{eq:app-c-timevarying-optimum}
\inf_{(\mu_k)}\limsup_{n\to\infty}
\left(\sup_{\|x\|_2=1}\mathbb E\|R_{n-1}\cdots R_0x\|_2^2\right)^{1/n}
=q.
\end{equation}
A mean-square exponentially stable independent sequence exists exactly when $\rho(A)<1$ and $\mathcal E(A)<r$.
\end{theorem}

\begin{proof}
Fix $n$ and the first $n$ laws. On $(\mathbb R^d)^n$, define a random block-cyclic matrix $\widehat R$ by \((\widehat Rz)_{j+1\bmod n}=S_jz_j,\; 0\le j<n\), where the $S_j\sim\mu_j$ are independent. If $C_n$ is the cyclic permutation matrix, then
\[
\mathbb E\widehat R=C_n\otimes A,\;
\operatorname{rank}\widehat R=\sum_{j=0}^{n-1}\operatorname{rank} S_j\le nr.
\]
The lifted law has finite second moment. The eigenvalues of $C_n\otimes A$ are $\zeta\lambda_i(A)$, where $\zeta$ ranges over the $n$th roots of unity. Thus $\rho(C_n\otimes A)=\rho(A)$ and $\mathcal E(C_n\otimes A)=n\mathcal E(A)$. Applying the optimal-value theorem in Section~4 in dimension $nd$ with rank budget $nr$ yields
\begin{equation}
\label{eq:app-c-lift-lower}
\rho(\widehat{\mathcal M})\ge q,
\; \widehat{\mathcal M}(X):=\mathbb E[\widehat RX\widehat R^{\top}].
\end{equation}

Starting in block $j$, a full cycle has second-moment operator \(\mathcal T_j:=\mathcal M_{j+n-1\bmod n}\cdots\mathcal M_{j\bmod n}\). Cyclic permutations of a product of square linear maps have the same characteristic polynomial, so every $\mathcal T_j$ has spectral radius $\rho(\mathcal M_{n-1}\cdots\mathcal M_0)$. A product of $mn$ independent copies of $\widehat R$ is block diagonal. Its $j$th diagonal block consists of $m$ independent full cycles. Therefore
\[
\bigl\|(\widehat{\mathcal M}^*)^{mn}(I_{nd})\bigr\|_2
=\max_{0\le j<n}\bigl\|(\mathcal T_j^*)^m(I_d)\bigr\|_2.
\]
The positive-map norm identity in the proof of Lemma~\ref{lem:lyapunov} and Gelfand's formula give
\begin{equation}
\label{eq:app-c-lift-radius-identity}
\rho(\widehat{\mathcal M})^n
=\rho(\mathcal M_{n-1}\cdots\mathcal M_0).
\end{equation}
The norm identity applies on the full symmetric-matrix space, so it also accounts for cross-moment blocks outside the block-diagonal subspace. Combining \eqref{eq:app-c-lift-lower} and \eqref{eq:app-c-lift-radius-identity} proves \eqref{eq:app-c-operator-product-lower}.

Independence also gives
\[
\sup_{\|x\|_2=1}\mathbb E\|R_{n-1}\cdots R_0x\|_2^2
=\bigl\|(\mathcal M_{n-1}\cdots\mathcal M_0)^*(I_d)\bigr\|_2.
\]
The right side is the induced operator norm of the adjoint positive map on the symmetric matrices, with the matrix operator norm as the underlying norm. It is therefore at least the spectral radius of that map, proving \eqref{eq:app-c-finite-product-lower}.

The lower bound holds for every $n$. Constant schedules are also allowed, and the optimal-value theorem in Section~4 supplies iid laws with factors arbitrarily close to $q$, proving \eqref{eq:app-c-timevarying-optimum}. If $q\ge1$, equation~\eqref{eq:app-c-finite-product-lower} excludes mean-square exponential stability. If $q<1$, the optimal-value theorem in Section~4 supplies a stable iid law. The formula for $q$ gives the stated criterion.
\end{proof}

This theorem covers arbitrary deterministic and periodic sampling schedules. Dependence on past samples permits a different outcome. The following construction remains unbiased conditional on the entire sampling history, yet improves the stability threshold.

\subsubsection{A conditionally unbiased Markov construction}
\label{app:markov-sampling}

\begin{proposition}[Conditionally unbiased Markov sampling]
\label{prop:markov-counterexample}
Let $A=aI_2$ with $1/\sqrt2<a<1$, and choose $0<\varepsilon<1-a^2$. There is a stationary sequence with finitely many possible matrix values satisfying
\[
\operatorname{rank} R_k=1,\; \mathbb E[R_k\mid\mathcal F_k]=A,
\; \|R_k\|_2\le a/\varepsilon,
\]
where $\mathcal F_k$ contains all sampling information before step $k$, such that, for every $x\in\mathbb R^2$ and $n\ge1$,
\begin{equation}
\label{eq:app-c-markov-exact}
\mathbb E\|R_{n-1}\cdots R_0x\|_2^2
=\frac1{2\varepsilon}\left(\frac{a^2}{1-\varepsilon}\right)^n\|x\|_2^2.
\end{equation}
This sequence is mean-square exponentially stable, whereas no independent rank-one sequence with mean $A$ is mean-square stable. Over conditionally unbiased dependent sequences with rank at most one and finite second moments of all finite products, the corresponding infimum is
\[
 \inf_{(R_k)}\limsup_{n\to\infty}
 \left(\sup_{\|x\|_2=1}\mathbb E\|R_{n-1}\cdots R_0x\|_2^2\right)^{1/n}=a^2.
\]
\end{proposition}

\begin{proof}
Let $(I_k)_{k\ge-1}$ be the stationary Markov chain on $\{1,2\}$ with
\[
T=\begin{pmatrix}1-\varepsilon&\varepsilon\\\varepsilon&1-\varepsilon\end{pmatrix},
\; \Pr(I_{-1}=1)=\Pr(I_{-1}=2)=\tfrac12.
\]
Set $P_j=e_je_j^{\top}$ and
\[
R_k:=\frac{a}{T_{I_{k-1},I_k}}P_{I_k},
\; \mathcal F_k:=\sigma(I_{-1},\ldots,I_{k-1}).
\]
Then
\[
\mathbb E[R_k\mid\mathcal F_k]
=\sum_{j=1}^{2}T_{I_{k-1},j}\frac a{T_{I_{k-1},j}}P_j=aI_2.
\]
The rank is one, and $0<\varepsilon<1-a^2<1/2$ gives the asserted norm bound.

Since $P_iP_j=0$ for $i\ne j$, the product $\Phi_n:=R_{n-1}\cdots R_0$ vanishes unless $I_0=\cdots=I_{n-1}=j$ for some $j$. The event $I_{-1}=i$, $I_0=\cdots=I_{n-1}=j$ has probability $\tfrac12T_{ij}(1-\varepsilon)^{n-1}$, and on this event \(\Phi_n=\frac{a^n}{T_{ij}(1-\varepsilon)^{n-1}}P_j\). Summing these contributions gives
\[
\mathbb E[\Phi_n^{\top}\Phi_n]
=\frac{a^{2n}}{2(1-\varepsilon)^{n-1}}
\left(\frac1{1-\varepsilon}+\frac1\varepsilon\right)I_2
=\frac1{2\varepsilon}\left(\frac{a^2}{1-\varepsilon}\right)^nI_2,
\]
which proves \eqref{eq:app-c-markov-exact}. The factor is below one because $\varepsilon<1-a^2$. Theorem~\ref{thm:timevarying} gives the independent lower bound $q_1(aI_2)=2a^2>1$.

For any conditionally unbiased sequence in the stated class, iterating conditional expectations gives $\mathbb E\Phi_n=A^n$. Jensen's inequality implies $\mathbb E\|\Phi_nx\|_2^2\ge\|A^nx\|_2^2=a^{2n}\|x\|_2^2$, so its asymptotic factor is at least $a^2$. Letting $\varepsilon\downarrow0$ in \eqref{eq:app-c-markov-exact} gives the matching infimum.
\end{proof}

For $a=0.9$ and $\varepsilon=0.1$, the independent optimum is $1.62$, whereas this construction gives $\mathbb E\|\Phi_nx\|_2^2=5(0.9)^n\|x\|_2^2$. The samples are $P_1,P_2,9P_1,9P_2$. The construction records only the previous sampling index. Switching directions annihilates the existing product, while rescaling by the transition probability preserves conditional unbiasedness. Independence is therefore essential to the rank threshold in the optimal-value theorem in Section~4.

\subsection{Computation and reproducibility}\label{app-toc-c-5}
\label{app:computation}
The spectral and similarity studies contain $108$ static laws
and $300$ dynamic laws, respectively. Their expectations and
moment operators are evaluated by summing over the complete
finite supports. The maximum relative residual in the mean
constraint is below $6.59\times10^{-16}$, and the maximum
error in the attained dynamic optimum is below
$3.34\times10^{-15}$.
The two-dimensional identities and the three-point family approaching an unattained optimum in Section~5 are also evaluated in exact rational arithmetic. The Markov product formula in
Proposition~\ref{prop:markov-counterexample} is checked by rational
enumeration of every sampling path through eight steps.
The finite-support computations include the rare surviving
trajectories in the identical-moment example automatically.

\end{document}